\documentclass[10pt,twocolumn,letterpaper]{article}

\usepackage{cvpr}
\usepackage{times}
\usepackage{epsfig}
\usepackage{graphicx}
\usepackage{amsmath}
\usepackage{amssymb}
\usepackage{enumitem}

\usepackage{booktabs}

\usepackage{tabulary}
\usepackage[dvipsnames]{xcolor}
\usepackage[caption=false]{subfig}
\usepackage[british,english,american]{babel}
\usepackage{xspace}\xspace
\usepackage{multirow}
\usepackage{adjustbox}
\usepackage{array}
\usepackage[T1]{fontenc}
\usepackage{xurl}  % https://tex.stackexchange.com/questions/3033/forcing-linebreaks-in-url

\usepackage{transparent}

\usepackage{algorithm}

\usepackage{listings}
\definecolor{codegreen}{rgb}{0,0.5,0}
\definecolor{codeblue}{rgb}{0.25,0.5,0.5}
\definecolor{codegray}{rgb}{0.6,0.6,0.6}

\definecolor{citecolor}{HTML}{0071bc}
\usepackage[pagebackref=false,breaklinks=true,
letterpaper=true,colorlinks,citecolor=citecolor,
bookmarks=false,urlcolor=blue]{hyperref}

\newcommand{\app}{\raise.17ex\hbox{$\scriptstyle\sim$}}

\newcommand{\apbbox}[1]{AP$^\text{bb}_\text{#1}$}
\newcommand{\apmask}[1]{AP$^\text{mk}_\text{#1}$}

\newlength\savewidth\newcommand\shline{\noalign{\global\savewidth\arrayrulewidth
  \global\arrayrulewidth 1pt}\hline\noalign{\global\arrayrulewidth\savewidth}}
\newcommand{\tablestyle}[2]{\setlength{\tabcolsep}{#1}\renewcommand{\arraystretch}{#2}\centering\footnotesize}

\makeatletter\renewcommand\paragraph{\@startsection{paragraph}{4}{\z@}
  {.5em \@plus1ex \@minus.2ex}{-.5em}{\normalfont\normalsize\bfseries}}\makeatother

\def\x{\times}
\cvprfinalcopy % *** Uncomment this line for the final submission

\def\cvprPaperID{999} % *** Enter the CVPR Paper ID here

\begin{document}

%%%%%%%%% TITLE
\title{Rethinking ``Batch'' in BatchNorm}

\author{
\vspace{.5em}
Yuxin Wu \quad Justin Johnson \\
Facebook AI Research
}

\maketitle
%\thispagestyle{empty}

%%%%%%%%% ABSTRACT
\begin{abstract}
\vspace{-.5em}
BatchNorm is a critical building block in modern convolutional neural networks.
Its unique property of operating on ``batches'' instead of individual samples
introduces significantly different behaviors from most other operations in deep learning.
As a result, it leads to many hidden caveats that can negatively impact
model's performance in subtle ways.
This paper thoroughly reviews such problems in visual recognition tasks,
and shows that a key to address them is to rethink different choices in the
concept of ``batch'' in BatchNorm.
By presenting these caveats and their
mitigations, we hope this review can help researchers use BatchNorm
more effectively.

\vspace{-1em}
\end{abstract}
%File:

\section{Introduction}

BatchNorm~\cite{Ioffe2015} is a critical component of modern convolutional neural networks (CNNs).
It is empirically proven to make models less sensitive to learning rates and initialization, and enables training for a wide variety of network architectures.
It improves model convergence speed, and also provides a regularizing effect to combat overfitting.
Due to these attractive properties, since its invention BatchNorm
has been included in nearly all state-of-the-art CNN architectures~\cite{He2016,Hu_2018_CVPR,pmlr-v97-tan19a,Radosavovic_2020_CVPR}.

% BatchNorm~\cite{Ioffe2015} is a critical innovation in the research of neural network architectures.
% It is empirically proven to be very effective in improving gradient-based training
% in convolutional neural networks (CNNs):
% It enables training for a variety of models that are not trainable otherwise. It also improves convergence and regularization for models that are already trainable.
% As a result, BatchNorm is widely used in almost all state-of-the-art CNN models
% since it was invented.

Despite its benefits,
% Though it has many benefits,
there are many subtle choices about precisely how BatchNorm can be applied in different scenarios.
Making suboptimal choices can silently degrade model performance:
models will still train, but may converge to lower accuracies.
For this reason, BatchNorm is sometimes viewed as
a ``necessary evil'' \cite{yanninterview}
in the design of CNNs.
The goal of this paper is to summarize the pitfalls that can befall practitioners when applying BatchNorm,
and provide recommendations for overcoming them.

% However, BatchNorm is also one of the most complicated building blocks in
% common CNN architectures, in both its design and its implementation.
% When applying BatchNorm in different scenarios,
% it introduces many caveats that one may easily overlook,
% which as a result silently compromise model performance.
% For this reason, BatchNorm is sometimes viewed as
% a ``necessary evil''
% \footnote{
%   Exclusive Interview with Yann LeCun. \textit{Computer Vision News}, Nov. 2018, page 9.
%    \url{https://www.rsipvision.com/ComputerVisionNews-2018November/Computer\%20Vision\%20News.pdf}}.
% To properly use BatchNorm in different scenarios, the complexity of
% BatchNorm extends beyond its standard form that are commonly used.

\begin{figure}[t]\centering
  \vspace{1.em}
  \includegraphics[width=0.9\linewidth]{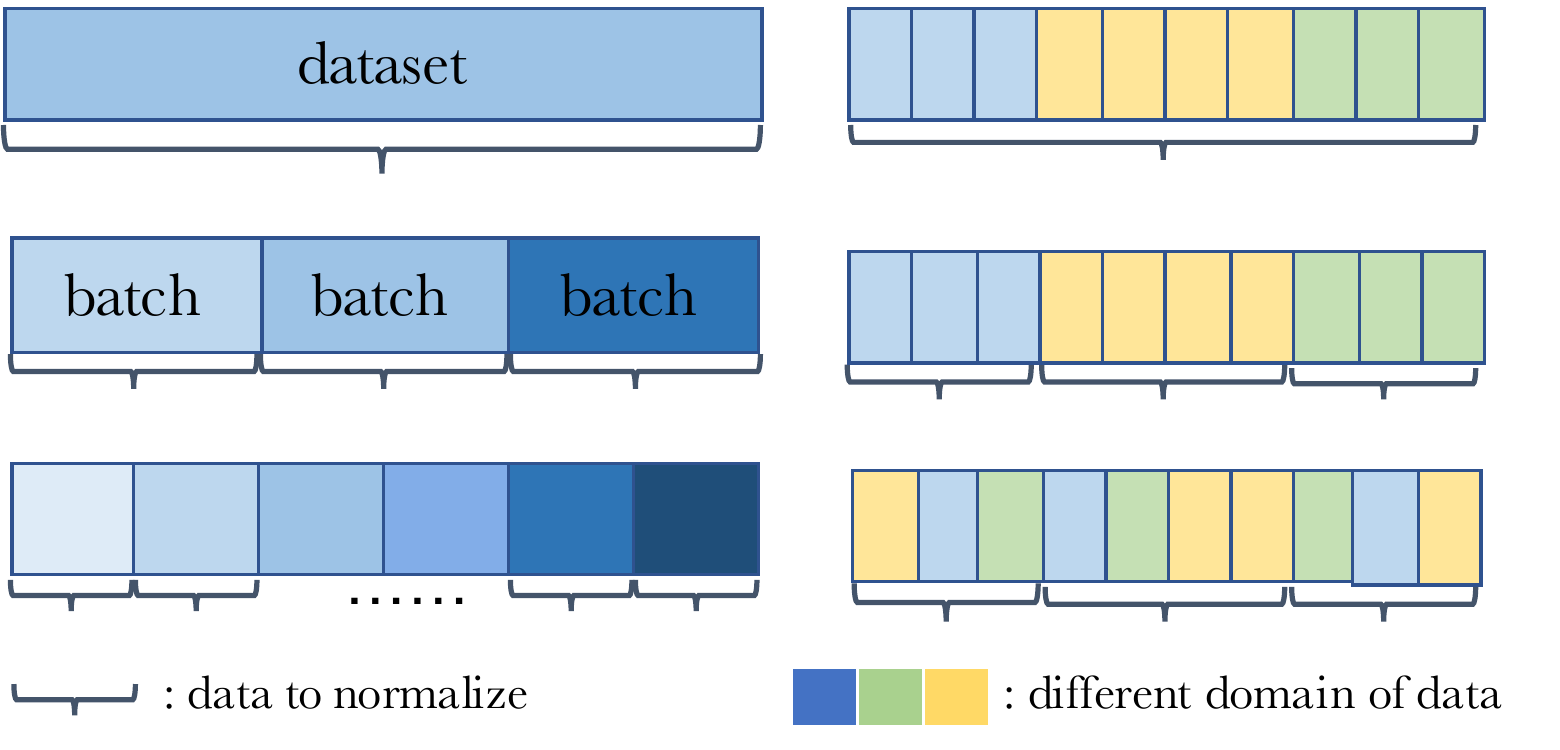}
  \caption{Examples of different choices of what is the ``batch'' to normalize over.
  Left: statistics can be computed from the entire dataset,
  mini-batches during SGD, or subset of mini-batches.
  Right: when inputs come from different domains, statistics can be computed from
  them combined, from each domain separately, or from a mixture of samples from
  each domain.
  }
  \vspace{-1em}
  \label{fig:teaser}
  \end{figure}

The key factor that differentiates BatchNorm from other deep learning operators
is that it operates on \emph{batches} of data rather than individual samples.
BatchNorm mixes information across the batch to compute normalization statistics,
while other operators process each sample in a batch independently.
BatchNorm's output thus depends not only on the properties of individual samples,
but also on the way samples are grouped into batches.
To show some different ways samples may be grouped together, Figure~\ref{fig:teaser} (left) illustrates that
the \emph{batch} over which BatchNorm operates may differ from the mini-batch used in
each step of stochastic gradient descent (SGD).
Figure~\ref{fig:teaser} (right) further shows additional choices
for choosing BatchNorm's batch when samples come from different domains.

This paper examines these choices for the \emph{batch} in BatchNorm.
We demonstrate that naively applying BatchNorm without considering different choices
for batch construction can have negative impact in many ways, but that model performance can be
improved by making careful choices in the concept of batch.
% More specifically:

% Section~\ref{sec:population-statistics} examines choices for using the entire training
% dataset as the batch, as is often done during inference.
% We discuss shortcomings in the commonly used exponential moving average (EMA) approach
% for computing population statistics, and argue for the use of more precise

% In many cases, we found that a naive application of BatchNorm without
% recognizing the differences between these choices can negatively affect the model.
% In this work, we re-examine different ways normalization statistics
% can be computed in BatchNorm, and analyze their implications on model's
% performance.
% By careful consideration of the concept of ``batch'',
% we show improvements in many models where BatchNorm would have caused trouble
% otherwise.
% More specifically, this paper discusses the following main topics about
% ``batch'' in BatchNorm:

Sec.~\ref{sec:population-statistics} discusses normalization statistics used during inference, where BatchNorm's ``batch'' is the entire training population.
We revisit the common choice of using an exponential moving average (EMA) of mini-batch statistics,
and show that EMA can give inaccurate estimates which in turn lead to
unstable validation performance.
We discuss PreciseBN as an alternative that more accurately estimates population statistics.

Unlike most neural network operators, BatchNorm behaves differently during training and inference:
it uses mini-batch statistics during training and population statistics during inference.
Sec.~\ref{sec:train-test-inconsistency} focuses on inconsistencies that can arise from this setup, and demonstrates
cases where it is beneficial to either use mini-batch statistics
during inference or population statistics during training.

Sec.~\ref{sec:mini-batch-domain} examines situations
where inputs of BatchNorm come from different domains,
either due to the use of multiple datasets, or due to layer sharing.
We show that in such settings, suboptimal choices of batches to normalize can
cause domain shift.

BatchNorm's use of mini-batch statistics during training can cause a subtle leakage of information between training samples.
Sec.~\ref{sec:info-leak} studies this behavior and shows that a careful choice of normalization batches can mitigate such potentially harmful effects.

% \begin{itemize}
%   \item During inference, ``batch'' is often meant to be the whole population,
%   but the commonly used exponential moving average (EMA) method is only an approximation using training mini-batches.
%   In Sec.~\ref{sec:population-statistics}, we show why EMA is not ideal
%   and study alternative methods to compute population statistics.
%   \item Sec.~\ref{sec:train-test-inconsistency} studies the
%   inconsistency that arise from the different definition of ``batch'' in
%   training vs. testing, and how the train-test
%   inconsistency can be reduced by redefining the ``batch'' in training and testing.
%   \item Sec.~\ref{sec:mini-batch-domain} studies what ``batch'' to normalize when
%   inputs come from different domains.
%   We show that suboptimal choices of batch causes or aggravates domain shift.
%   \item The BatchNorm operation could unintentionally leak information about samples
%   to others within a batch. Sec.~\ref{sec:info-leak} studies this
%   behavior and provides mitigations.
% \end{itemize}

In each section, we experiment with well-studied visual recognition tasks,
to demonstrate that a seemingly standard way to apply BatchNorm
can result in suboptimal behavior that subtly degrades model performance.
In each case, an alternative, uncommon way to compute BatchNorm's normalization statistics leads to significant improvements.

This paper is a \emph{review} of existing problems and their solutions.
Many of the topics we will discuss have been previously mentioned in the literature;
however they are scattered across many papers and may not be widely appreciated.
In our experience, these subtleties about BatchNorm frequently lead to
troubles when developing new models.
Therefore, we aim to collect much of the ``dark knowledge'' about BatchNorm into one place,
and we hope this review can serve as a reference guide to help practitioners avoid common pitfalls due to BatchNorm.

%File:

\section{A Review of BatchNorm}
\label{sec:review-of-batchnorm}
Here we briefly review Batch Normalization, focusing on its use in CNNs.
The inputs to BatchNorm are
CNN features $x$ of shape $(C, H, W)$ with $C$ channels
and spatial size $H\times W$.
BatchNorm computes an output $y$ which normalizes $x$ using per-channel
statistics $\mu,\sigma^2\in\mathbb{R}^C$:
\vspace{-1mm}
\begin{align*}
  y = \frac{x - \mu \footnotemark}{\sqrt{\sigma^2 + \epsilon}}
\end{align*}
\vspace{-3mm}
\footnotetext{Using numpy-style broadcasting, $\mu_c$ and $\sigma^2_c$ are reused for all $x_{c,\star,\star}$.}

% We give a brief review of Batch Normalization.
% We will discuss Batch Normalization in the context of CNNs:
% the input of a BatchNorm layer consists of CNN feature
% $x$ of shape $(C, H, W)$, where $C$ stands for the number of feature channels , and $H, W$ for the spatial dimensions.
% BatchNorm normalizes the input feature $x$ by $\mu, \sigma^2$:
% \begin{align*}
%   y = \frac{x-\mu}{\sqrt{\sigma^2+\epsilon}}
%   %\texttt{for each } & \texttt{channel } c \; (0 \le c < C) \texttt{:} \\
% %y[c,:,:] = & \frac{x[c,:,:] - \mu}{\sqrt{\sigma^2 + \epsilon}}
% \end{align*}
% where $\mu, \sigma^2$ are per-channel statistics
% computed from \textit{some} set of features.

The exact mechanism to compute $\mu$ and $\sigma^2$ can vary.
The most common setting (which gives BatchNorm its name) is for $\mu$ and $\sigma^2$
to be the \emph{mini-batch statistics} $\mu_{\mathcal{B}}, \sigma_\mathcal{B}^2$, i.e.~the empirical mean and variance of a mini-batch of $N$ features during training.
The mini-batch $X$ is packed into a 4D tensor of shape $(N, C, H, W)$; then
\begin{align*}
  \mu_\mathcal{B} = & \text{ mean}(X, \text{axis=}[N, H, W]) \footnotemark \\
  \sigma_\mathcal{B}^2 = &\text{ var}(X, \text{axis=}[N, H, W])
\end{align*}
\vspace{-4mm}
\footnotetext{Following the spirit of ``NamedTensor'' \cite{namedtensor}, this computes
the mean over the $N,H,W$ dimensions resulting in a vector of length $C$.}

During inference $\mu$ and $\sigma^2$ are not computed from mini-batches;
instead, \emph{population statistics} $\mu_{pop},\sigma^2_{pop}$ are estimated from the training set and used for normalization.

% The exact set of features from which to compute statistics $\mu, \sigma^2$
% can vary.
% As one common choice, they can be
% a mini-batch of $N$ features represented by a 4D tensor of shape $(N,C,H,W)$:
% \begin{align*}
%   \mu_B = & \text{ mean}(X, \text{axis=}[N, H, W]) \footnotemark \\
%   \sigma_B^2 = &\text{ var}(X, \text{axis=}[N, H, W])
% \end{align*}
% \footnotetext{Following the spirit of ``NamedTensor'' \cite{namedtensor}, this stands for
% computing the mean on the $N, H, W$ dimensions. The result is a vector of length $C$. }

As we will see in the paper, there could be many other choices
about what is the ``batch'', i.e. what is the data on which we compute $\mu, \sigma^2$.
The size of the batch, data source of the batch, or algorithm to compute the
statistics can vary in different scenarios,
leading to inconsistencies which in the end affect generalization of models.

Our definition of BatchNorm in this paper differs slightly from convention:
we do not view the channel-wise affine transform that typically follows normalization as part of BatchNorm.
All our experiments contain this affine transform as usual, but we view it as a separate, regular layer
since it operates independently on samples (it is actually equivalent to a depth-wise $1\times1$ convolutional layer).
This distinction focuses our attention on the unique use of the ``batch'' in BatchNorm.

By excluding the affine transform, the population statistics
$\mu_{pop}, \sigma_{pop}$ become BatchNorm's only learnable parameters.
However unlike other parameters in a neural network,
they are \emph{neither used nor updated} by gradient descent.
Instead, population statistics are trained using other algorithms (e.g. EMA),
and Sec.~\ref{sec:population-statistics} will consider alternative algorithms.
Viewing the computation of population statistics as a form of training can help us
diagnose this process with the standard concept of generalization.
In Sec.~\ref{sec:mini-batch-domain} we will see that poorly trained population
statistics can affect generalization due to domain shift.
\section{Whole Population as a Batch}
\label{sec:population-statistics}
During training, BatchNorm computes
normalization statistics with a mini-batch of samples.
However, when the model is used for testing,
there is usually no concept of mini-batch any more.
It is originally proposed~\cite{Ioffe2015} that at test-time,
features should be normalized by
population statistics $\mu_{pop},\sigma_{pop}$,
computed over the entire training set.
Here $\mu_{pop}, \sigma_{pop}$ are defined as the batch statistics $\mu_\mathcal{B}, \sigma_\mathcal{B}$
using the whole population as the ``batch''.

Because $\mu_{pop}, \sigma_{pop}$ are computed from data,
they shall be considered \emph{trained} on the given data, but
with different algorithms.
Such training procedure is often unsupervised and cheap,
but it's of critical importance to model generalization.
In this section,
we'll show that the widely used EMA algorithm does not always accurately
train the population statistics, and we discuss alternative algorithms.

\begin{figure*}[ht!]\centering
  \resizebox{1\linewidth}{!}{
    \subfloat{
      \includegraphics[width=1.0\linewidth]{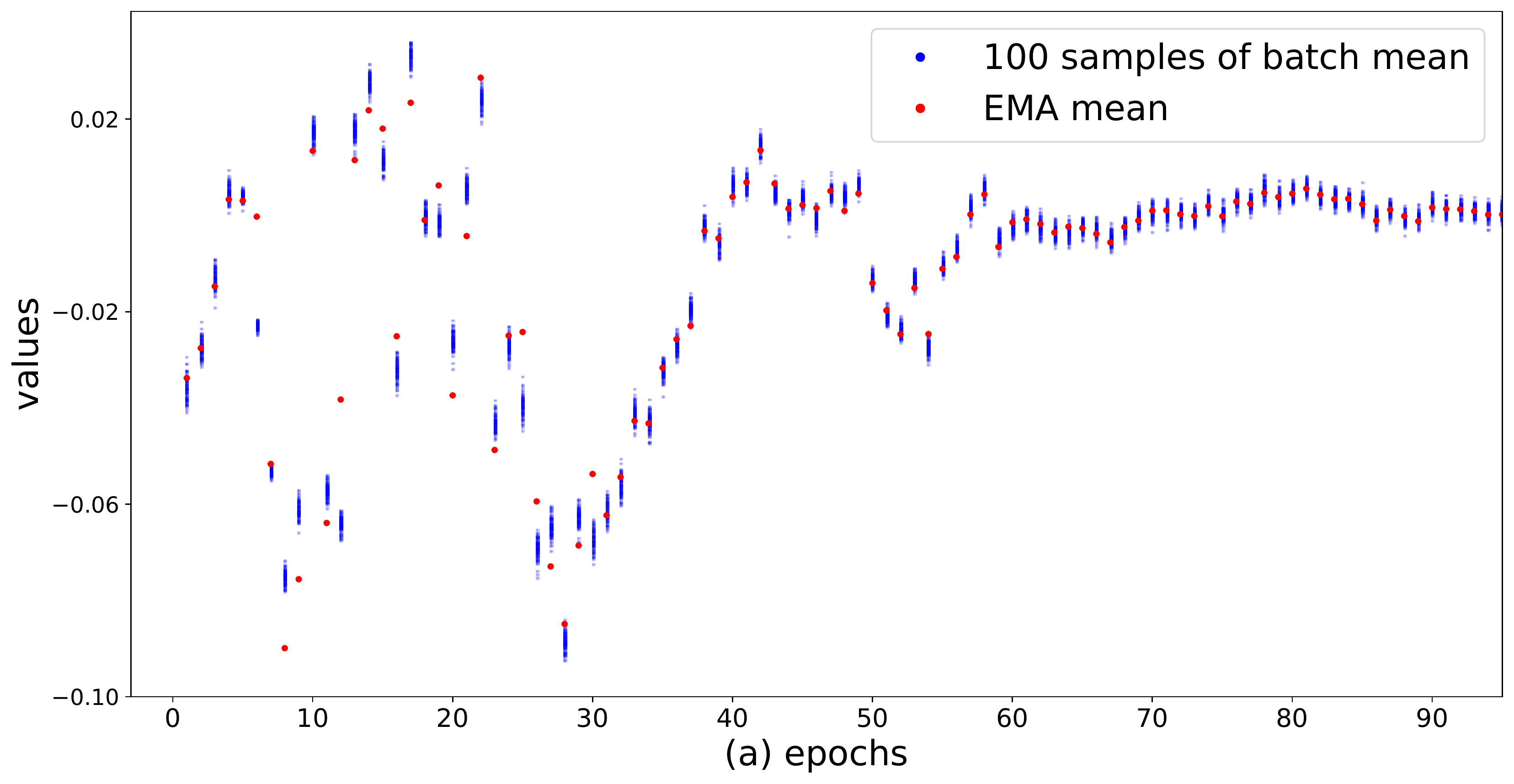}
    }
    \subfloat{
      \includegraphics[width=0.98\linewidth]{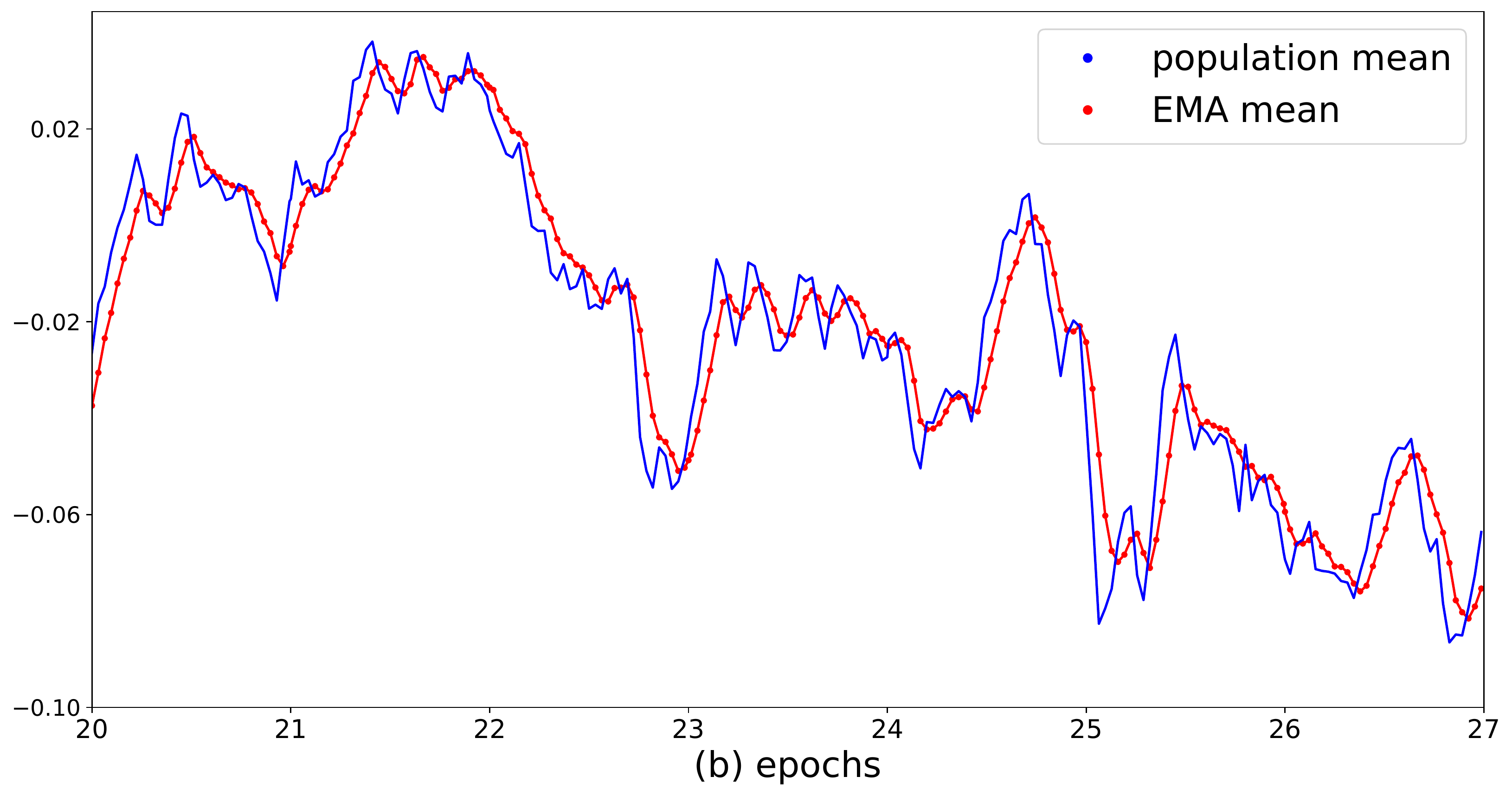}
    }
  }
  \caption{
    Values of EMA means ({\color{red} red}), batch means ({\color{blue} blue}), population means ({\color{blue}blue})
    of a random channel in a random BatchNorm layer of a ResNet-50 during training.
    For each checkpoint, we compute 100 batch means by applying it on 100 mini-batches of data.
    Population mean is approximated by a simple average of 100 batch means.
    (a) shows that EMA often falls out of the distribution of mini-batch
    statistics, especially at the earlier stage of training.
    (b) shows more details of (a) on the 20th - 27th epochs:
    EMA values lag behind the statistics produced by the
    current model, due to its weighting on historical samples.
    Similar patterns exist in other channels and BN layers we have checked.
    }
  \label{fig:ema-scatter}
  \end{figure*}

\subsection{Inaccuracy of EMA}
\cite{Ioffe2015} proposes that
exponential moving average (EMA) can be used
to efficiently compute population statistics.
Such approach has nowadays become a standard in deep learning libraries.
However, we will see in this section that despite its prevalence,
EMA may not produce a good estimate of population statistics.

To estimate $\mu_{pop}, \sigma_{pop}$,
EMAs are updated in every training iteration by:
  \begin{align*}
  \mu_{EMA} \leftarrow & \lambda \mu_{EMA} + (1-\lambda) \mu_\mathcal{B} \\
\sigma^2_{EMA} \leftarrow  &\lambda \sigma^2_{EMA} + (1-\lambda) \sigma_\mathcal{B}^2
  \end{align*}
where $\mu_\mathcal{B}, \sigma_\mathcal{B}^2$ are batch mean and variance of each iteration
and momentum $\lambda \in [0, 1]$.

The exponential averaging
can lead to suboptimal estimate of population statistics for the following reasons:
\begin{itemize}[leftmargin=*]
  \item Slow convergence of statistics when $\lambda$ is large.
  Since each
  update iteration only contributes a small portion ($1-\lambda$) to EMA,
  a large number of updates are needed for EMA to converge to a stable estimate.
  The situation worsen as the model is being updated:
  EMA is largely dominated by input features from the past,
  which are outdated as the model evolves.
  \item When $\lambda$ is small, EMA statistics becomes dominated by
  a smaller number of recent mini-batches
  and does not represent the whole population.
\end{itemize}

Next we demonstrate the inaccuracy of EMA by standard ImageNet classification.
We train a ResNet-50~\cite{He2016} on ImageNet~\cite{Deng2009} for 100 epochs,
with all experimental settings following~\cite{Goyal2017}:
we use a total mini-batch size of 8192 across 256 GPUs,
while BatchNorm is performed within the 32 samples on each GPU.
EMA is updated with the per-GPU statistics using $\lambda=0.9$.
We frequently save the model's parameters during training, in order to plot how
the EMA statistics $\mu_{EMA}, \sigma_{EMA}$ evolve for an arbitrary channel of a BatchNorm layer that
we randomly select from the model.
In addition,
we also forward the saved checkpoints under ``training mode''
to compute batch statistics of 100 random mini-batches (each also has 32 samples)
that are plotted together with the EMA statistics.
The plots in Fig.~\ref{fig:ema-scatter} show that EMA is
not able to accurately represent mini-batch statistics or population statistics
in the early stage of training.
Therefore, using EMA as the ``population statistics'' may harm the accuracy of the model.
We will analyze the impact to model accuracy in the next section.

\subsection{Towards Precise Population Statistics}
\label{sec:precisebn}

Ideally we would like to compute more precise population statistics following
its definition: the mean and variance of features using the entire training dataset as one batch.
Because such a large batch may be too expensive to process,
we aim to approximate the true population statistics by the following two steps:
(1) apply the (fixed) model on many mini-batches to collect batch statistics;
(2) aggregate the per-batch statistics into population statistics.

This approximation has kept two important properties of true population statistics that are
different from EMA:
(1) statistics are computed entirely from a \emph{fixed model state}, unlike EMA which
uses historical states of the model; (2) all samples are equally weighted.
Next, we will see that this method produces better statistics that improve the model performance.
We will also discuss its implementation details and
show that this approximation works as well as true population statistics.

As EMA has now become the de-facto standard of BatchNorm,
we use the name ``\textbf{PreciseBN}'' when a BatchNorm layer uses such more precise
statistics in inference, to not be confused with the prevalent BatchNorm implementations.
However, we note that this approach is in fact the definition of the
original BatchNorm in~\cite{Ioffe2015}.
A PyTorch implementation of PreciseBN is provided in \cite{fvcore}.

%\begin{algorithm}[t]
  %\caption{Pseudocode to compute ``precise'' population statistics for BatchNorm}
  %\label{alg:precisebn}
  %\definecolor{codeblue}{rgb}{0.25,0.5,0.5}
  %\lstset{
    %backgroundcolor=\color{white},
    %basicstyle=\fontsize{7.2pt}{7.2pt}\ttfamily\selectfont,
    %columns=fullflexible,
    %breaklines=true,
    %captionpos=b,
    %commentstyle=\fontsize{7.2pt}{7.2pt}\color{codeblue},
    %keywordstyle=\fontsize{7.2pt}{7.2pt},
  %%  frame=tb,
  %}
  %\textbf{Input:} data\_iter: an iterable of training batches.

  %\textbf{Output:} $\mu_{pop}$, $\sigma_{pop}^2$: population statistics for BN layers.

  %mean\_list $ \leftarrow \left[\: \right]$, var\_list $ \leftarrow \left[\: \right]$

  %\textbf{for} data \textbf{in} data\_iter:

     %\hspace{1em}  forward with data, to obtain batch statistics $\mu, \sigma^2$.

     %\hspace{1em}  mean\_list.append($\mu$), var\_list.append($\sigma^2$)

     %\hspace{1em} \# the model is \textbf{not} updated with back-propagation

   %$\mu_{pop} \leftarrow $ mean\_list.mean()

   %$\sigma_{pop}^2 \leftarrow$ var\_list.mean()
  %\end{algorithm}

\label{sec:ema-failures}

\begin{figure}[t]\centering
  \includegraphics[width=0.9\linewidth]{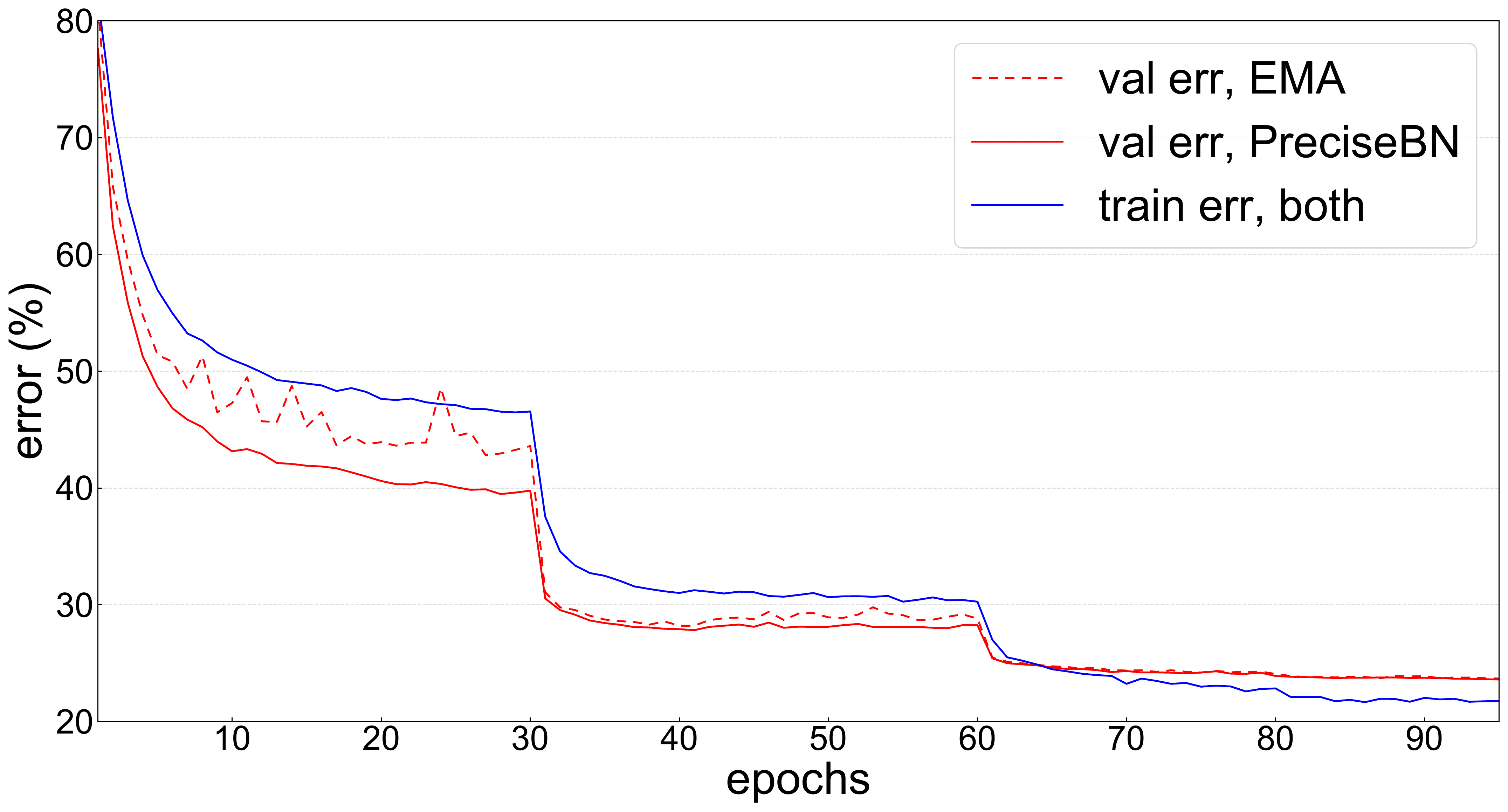}
  \caption{Training and validation error curves with EMA (more noisy) and PreciseBN.
  Note that training error (defined by the error rate of predictions made in
  training mode on augmented training data)
  is not affected by how population statistics is trained,
  so only one training curve is shown.
  }
  \vspace{-1em}
  \label{fig:precisebn-256}
  \end{figure}

\paragraph{PreciseBN stabilizes inference results.}
We train a ResNet-50 on ImageNet
following the recipe in~\cite{Goyal2017} with a regular total
batch size of 256 distributed on 8 GPUs.
The model is evaluated using precise statistics as well as EMA every epoch.
The error curves plotted in
Fig.~\ref{fig:precisebn-256} shows that
the validation results evaluated with PreciseBN is better and also more stable than
the one with EMA, demonstrating that the inaccuracy of EMA has negative
impact on model performance.
Final accuracy is listed in row 1 of Table~\ref{tab:population-size}.

\begin{figure}[t]\centering
  \includegraphics[width=0.9\linewidth]{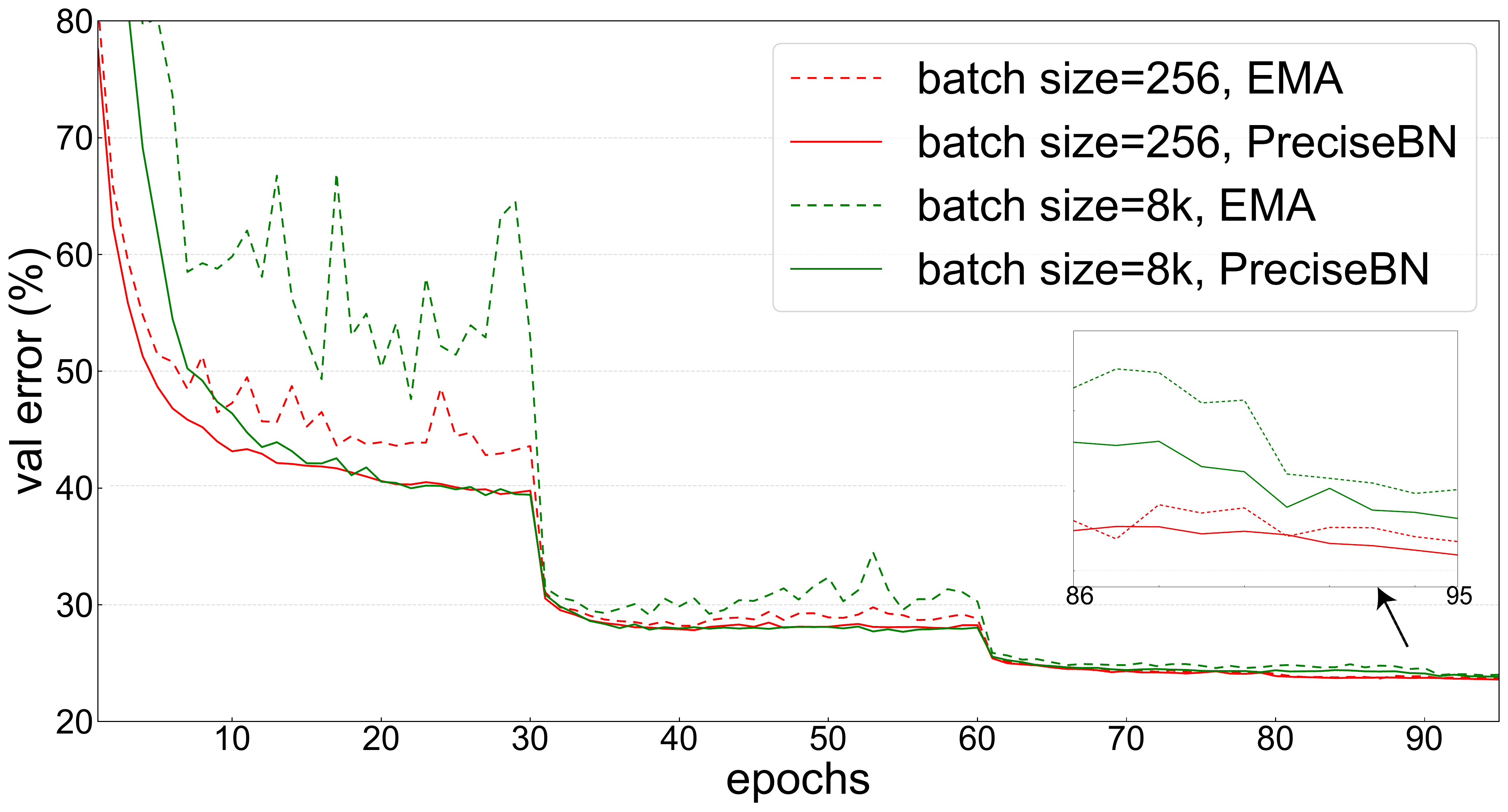}
  \caption{Validation curves using different test-time statistics (EMA vs. PreciseBN)
      and training batch sizes (256 vs. 8192).}
  \label{fig:precisebn-largebatch}
  \end{figure}

\paragraph{Large-batch training suffers more from EMA.}
We conduct the same experiment using a larger total batch size of 8192.
Following~\cite{Goyal2017}, the base learning rate
$\eta$ is linearly scaled to 32$\times$ larger.
Validation errors are plotted in Fig.~\ref{fig:precisebn-largebatch}.

Fig.~\ref{fig:precisebn-largebatch} shows that EMA produces extremely high-variance
validation results under a larger batch size.
This was previously observed in Fig. 4 of~\cite{Goyal2017} as well.
The phenomenon is most severe for the first 30 epochs of training when the learning rate is largest,
which matches our observation in Fig.~\ref{fig:ema-scatter} (a).
We believe the instability is contributed by two factors in large-batch training
that harm the convergence of EMA statistics:
(1) the 32$\x$ larger learning rate causes features to change more dramatically;
(2) EMA are updated for 32$\x$ fewer times, due to the reduced total training iterations.
On the other hand, PreciseBN produces stable accuracy.
Table~\ref{tab:population-size} lists their accuracy at the 30th and 100th epoch.

\paragraph{PreciseBN only requires $10^3 \sim 10^4$ samples.}
In practice, estimating mean and variance on the whole training set is expensive,
and accurate population statistics can be obtained from a sufficiently large subset.
In Table~\ref{tab:population-size},
we show how the error rate changes w.r.t. the number of samples used in PreciseBN,
using three models taken from the previous experiments.
The results show that PreciseBN is able to obtain stable performance with
only $N = 10^3\sim 10^4$ samples.
If we execute PreciseBN once in every epoch of ImageNet training, we estimate that using PreciseBN
increases the total training cost by merely 0.5\%
\footnote{Assume each epoch contains $10^6$ images and backward cost is the same as forward,
the additional training cost per epoch is $\frac{10^4}{2 \times 10^6} = 0.5\%.$
}.

Though only requiring $10^3\sim 10^4$ samples, PreciseBN cannot be easily
achieved by training with a large batch and using EMA with $\lambda=0$. We discuss this
in Appendix~\ref{sec:large-batch-ema-lambda}.

\begin{table}
\tablestyle{1.5pt}{1.05}
\begin{tabular}{cc||c|c|c|c|c}
Batch Size & Epoch & EMA & $N=10^1$ & $N=10^2$ & $N=10^3$ & $N=10^4$ \\ \shline
256& 100 & 23.7 & 28.1 \tiny{$\pm 0.5$} & 24.0\tiny{$\pm 0.1$} & 23.57\tiny{$\pm 0.05$} & 23.54 \tiny{$\pm 0.02$}\\
8192& 30 & 52.8 & 45.5\tiny{$\pm 1.3$} & 40.1\tiny{$\pm 0.2$} & 39.44\tiny{$\pm 0.06$} & 39.42\tiny{$\pm 0.03$} \\
8192& 100& 24.0 & 29.0\tiny{$\pm 0.7$} & 24.1\tiny{$\pm 0.1$} & 23.77\tiny{$\pm 0.05$} & 23.73\tiny{$\pm 0.02$}\\
\end{tabular}
\vspace{0.7em}
\caption{Validation errors (\%) for three models, evaluated using
EMA and PreciseBN with different
number of images $N$.
Standard deviation in the table comes from the random choice of samples.
$N\ge 10^3$ can provide reasonably good results, even when the model is far from
convergence (epoch 30).
}
\label{tab:population-size}
\end{table}

\paragraph{Aggregation of per-batch statistics.}
Even with $N < 10^4$, it is still often computationally challenging to forward
 sufficient amount of samples in one batch.
In practice,
$N$ samples need to be processed with a mini-batch size of $B$ for $k = \frac{N}{B}$ times.
This leads to the question of how to aggregate per-batch statistics $\mu_{\mathcal{B}i}, \sigma_{\mathcal{B}i}^2,
(i = 1 \cdots k)$
into $\mu_{pop}, \sigma_{pop}^2$.

We aggregate statistics by $\mu_{pop} = \mathrm{E}[\mu_\mathcal{B}], \sigma_{pop}^2 = \mathrm{E}[\mu_\mathcal{B}^2 + \sigma_{B}^2] - \mathrm{E}[\mu_\mathcal{B}]^2$
in all experiments, where $\mathrm{E}[\cdot]$ is the average over $k$ batches.
We discuss some alternative aggregation methods in Appendix~\ref{sec:variance},
but they all work similarly well in practice.

\paragraph{Small batch causes accumulated errors.}
When computing population statistics, the choice of batch size $B$ actually matters:
different batch grouping will change the normalization statistics
and affects output features.
This difference in features can then accumulate in deeper
BatchNorm layers, and lead to inaccurate statistics.

To verify this claim, we evaluate the same models using PreciseBN, with the same
population of $N=10^4$ samples but different batch size $B$
to forward these samples.
Results in Table~\ref{tab:precisebn-batchsize}
show that population statistics estimated from small mini-batch sizes are
inaccurate and degrade accuracy.
A batch size $B \ge 10^2$ appears sufficient in our experiments.

%Even in the extreme case where the batch has difficulty fit into RAM
%(e.g. $B=N$ in Table~\ref{tab:precisebn-batchsize}),
%results can still be accurately obtained by layer-wise computation
%using smaller batch size, discussed more in Appendix~\ref{sec:layer-wise-precise}.

\begin{table}
\tablestyle{1.5pt}{1.05}
\begin{tabular}{c|c||c|c|c|c|c}
Model & EMA & $B=2$ & $B=10$ & $B=10^2$ & $B=10^3$ & $B=N$ \\ \shline
NBS=32 & 23.7 & 28.8 & 24.0 & 23.5 & 23.5 & 23.5\\
NBS=2 & 35.5 & 34.4 & 32.3 & 32.0 & 32.0 & 32.0 \\
\end{tabular}
\vspace{0.7em}
\caption{Validation error of two ResNet-50 models trained for 100 epochs. Both are
trained under the same total SGD batch size of 256 but different normalization batch size
(NBS, defined in Sec.~\ref{sec:nbs}).
They are evaluated with EMA and precise statistics,
where precise statistics are computed with $N=10^4$ samples split under different batch sizes $B$.
}
\vspace{-1em}
\label{tab:precisebn-batchsize}
\end{table}

Because EMA has to estimate population statistics using the same mini-batch size used in SGD,
our results also suggest that EMA cannot estimate good population statistics
when trained with a small batch size.
On the other hand, the batch size $B$ in PreciseBN can be independently chosen,
and requires much less RAM because no back-propagation is needed.
For models trained with a normalization batch size of 2
(row 2 in Table \ref{tab:precisebn-batchsize}),
correct use of PreciseBN improves the accuracy by 3.5\% over EMA.

\subsection{Discussions and Related Works}
\label{sec:population-discussions}
\paragraph{PreciseBN.} We note that recomputing population statistics like PreciseBN is actually how the original BatchNorm
\cite{Ioffe2015} is formulated, but it is not widely used.
This paper thoroughly analyzes
EMA and PreciseBN under the standard ImageNet classification task, and demonstrate the advantage of PreciseBN
in appropriate situations.

We show that EMA is unable to precisely estimate population statistics
when the model is not converged, especially during large-batch training.
PreciseBN shall be used when a model needs to be evaluated in such scenarios.
In addition to drawing validation curves, reinforcement learning also
requires evaluating the model before convergence.
We review one such example from~\cite{tian2019opengo} in Appendix~\ref{sec:ema-rl}.
Even when the model has reached convergence, we show that EMA produces poor estimation if the model
is trained with a small batch size.

Nevertheless,
our experiments also show that
the final validation errors obtained by EMA is often as good as those from PreciseBN,
given large enough batch size and converged model.
As a result, the EMA approach is still dominant and its disadvantages are often
overlooked.
We hope our analysis helps researchers recognize the problems of EMA
and use PreciseBN when necessary.

\paragraph{Other advantages.} Recomputing population statistics is also found important in the presence of other train-test inconsistencies:
\cite{brock2018large,Pavel_SWA} recompute statistics because the model weights
go through averaging after training.
\cite{Li_dropout_bn} recomputes population variance
to compensate for the ``variance shift'' caused by
the inference mode of dropout.
\cite{shomron2020post,hubara2020improving,sun2019hybrid}
recompute / recalibrate population statistics to compensate for the
distribution shift caused by test-time quantization.
Recomputing population statistics on a different domain \cite{li2016revisiting}
is common in domain adaptation, which will be discussed further in Sec.~\ref{sec:mini-batch-domain}.
All these examples involve extra train-test inconsistencies
in either model or data,
which justify a re-evaluation of population statistics, as
the EMA statistics estimated during training may be inconsistent with
the feature distribution during testing.
State-of-the-art vision models often regularize the training by
strong train-test inconsistencies~\cite{Srivastava2014,Huang2016deep,zhang2018mixup,ghiasi2018dropblock,yun2019cutmix},
which may also interact with BatchNorm in unexpected ways.
%File:

\section{Batch in Training and Testing}
\label{sec:train-test-inconsistency}
BatchNorm typically has different behaviors in training and testing:
the normalization statistics come from different ``batches'', namely the mini-batch and the population.
The gap between population statistics and mini-batch statistics
introduces inconsistency.
In this section, we analyze the effect of such inconsistency on
model's performance, and point out that in circumstances the inconsistency
can be easily removed to improve performance.

\subsection{Effect of Normalization Batch Size}
\label{sec:nbs}
We define ``normalization batch size'' as the size of the actual mini-batch
on which normalization statistics are computed.
To avoid confusion, we explicitly use ``SGD batch size'' or ``total batch size'' in this paper to
refer to the mini-batch
size of the SGD algorithm, i.e. number of samples used to compute one gradient update.

In standard implementations of BatchNorm in major deep learning frameworks,
normalization batch size during training is equal to per-GPU batch size.
By using alternative implementations such as SyncBN~\cite{Peng2018},
GhostBN~\cite{hoffer2017train} discussed in Appendix~\ref{sec:changing-nbs},
we are able to increase or decrease normalization batch size easily.

\begin{figure}[t!]\centering
  \includegraphics[width=1.05\linewidth]{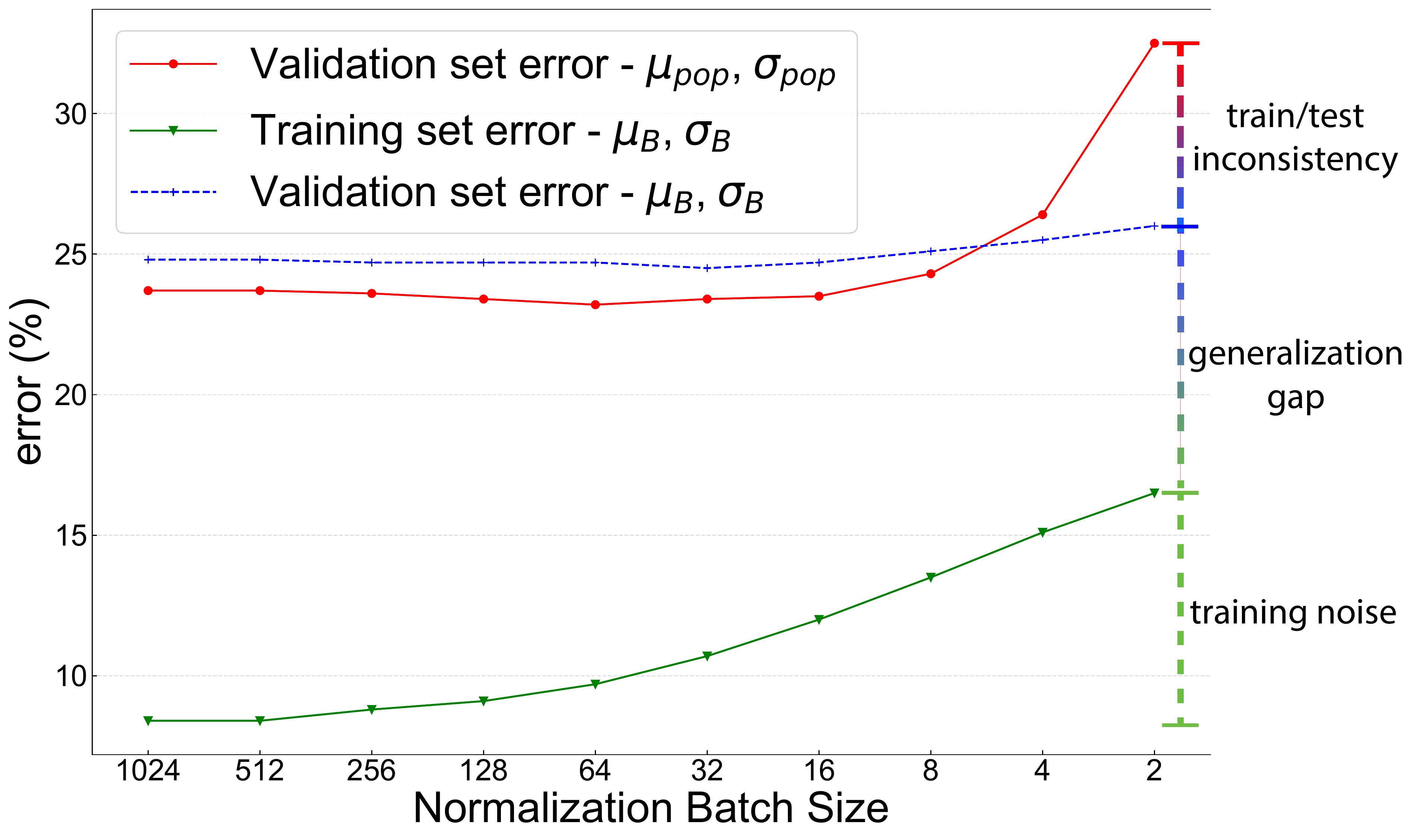}
  \caption[]{
  Classification error under different normalization batch sizes,
  with a fixed total batch size of 1024.
  {\color{OliveGreen} Green}: error rate on unaugmented training set using mini-batch statistics;
  {\color{Red} Red}: error rate on validation set using population statistics estimated by PreciseBN;
  {\color{blue} Blue}: error rate on validation set using mini-batch statistics
  of random batches
  (with the same normalization batch size used in training).
  The gap between red and blue curves is caused by train-test inconsistency,
  while the gap between blue and green curves is the generalization gap on unseen dataset.
  }
  \vspace{-1em}
  \label{fig:nbs}
  \end{figure}

Normalization batch size has a direct impact on training noise and train-test inconsistency:
a larger batch pushes the mini-batch statistics closer to
the population statistics, thus reduces training noise and train-test inconsistency.
To study such effect,
we train ResNet-50~\cite{He2016} models following the recipe
in~\cite{Goyal2017},
but with normalization batch sizes varying from 2 to 1024.
The SGD batch size is fixed to 1024 in all models.
For the purpose of analysis, we look at the error rate of each model under 3 different
evaluation schemes:
(1) evaluated with mini-batch statistics on training set;
(2) evaluated with mini-batch statistics on validation set;
(3) evaluated with population statistics on validation set.
We plot the final metrics of each model in Fig.~\ref{fig:nbs}.
By ablations among the three evaluation schemes,
we can observe the following from the figure:

\paragraph{Training noise.}
The monotonic trend of training set error (green)
is contributed by the amount of training noise during SGD.
When normalization batch size is small,
features of one sample are heavily affected by other randomly chosen samples in the same mini-batch,
causing optimization difficulties as shown by the poor training accuracy.

\paragraph{Generalization gap.}
When also evaluated with mini-batch statistics,
but on validation set (blue), the model performs worse compared to the training set (green).
The gap between them is purely a generalization gap
due to the change of dataset,
because the mathematical computation of the model has not changed.
The generalization gap monotonically decreases with the normalization batch size,
likely due to the regularization effect provided by training noise and
train-test inconsistency.

\paragraph{Train-test inconsistency.}
To analyze the inconsistency due to the use of $\mu_{pop}$, $\sigma_{pop}$, we compare
the use of population statistics
and mini-batch statistics, both on validation set (red vs. blue curves).

A small normalization batch size (e.g. 2 or 4) is known to
perform poorly \cite{Ioffe2017,Wu2018},
but the model in fact has decent performance if mini-batch statistics are used (blue curve).
Our results show that
the large inconsistency between mini-batch statistics and population statistics
is a major factor that affects performance at small batch sizes.

On the other hand, when normalization batch size is larger,
the small inconsistency
can provide regularization to reduce validation error.
This causes the red curve to perform better than the blue curve.

In this experiment, the best validation error is found at a normalization batch size
of 32$\sim$128, where the amount of noise and inconsistency provides
balanced regularization.

\subsection{Use Mini-batch in Inference}
\label{sec:batch-stats-in-inference}
Results in Fig.~\ref{fig:nbs} indicate that
using statistics of the mini-batch in inference, if applicable,
can reduce train-test inconsistency and improve test-time performance.
Surprisingly, we show that with this method, BatchNorm with a tiny batch size (2) only suffers a
performance drop less than 3\% (Fig.~\ref{fig:nbs}, blue)
compared to regular batch size.
However, using mini-batch statistics in inference is often not a legitimate
algorithm as the model no longer makes independent predictions for each sample.

\paragraph{R-CNN's head.} We now study a case where it is legitimate to use mini-batch statistics in inference.
We look at the second stage (a.k.a ``head'') of an R-CNN style object detector
\cite{Girshick2014,Girshick2015,Ren2015, He2017}.
\label{sec:maskrcnn-norm-head}
The head of an R-CNN takes a total of $R_i$ region-of-interest (RoIs)
for each input image $i$, and makes predictions for each RoI.
In most implementations of R-CNN, BatchNorm layers in the head
combine regions from all images
into a \emph{single} mini-batch of size $N = \sum_i R_i$.
We follow this convention in our experiments.

It is important to note that even if only one input image is given in inference,
the ``mini-batch'' with size $N=R_0$
(typically in the order of $10^2 \sim 10^3$ for a box head)
still exists, therefore mini-batch statistics can be used legitimately.

\begin{table}
\newcommand{\demph}[1]{\textcolor{Gray}{#1}}
\tablestyle{1.5pt}{1.00}
\begin{tabular}{c|c|c|cc}
Norm on Heads & \begin{tabular}{c}img / GPU \\ (training) \end{tabular}  & Stats. in inference &
\apbbox{~} &
\apmask{~} \\
\shline

per-GPU BN & 2 & $\mu_{pop},\sigma_{pop}$ &   40.1 & 36.2  \\
per-GPU BN & 2 & $\mu_\mathcal{B}, \sigma_\mathcal{B} $ (1 img)&  \textbf{41.3} & \textbf{37.0}  \\ \hline
per-GPU BN & 1 & $\mu_{pop}, \sigma_{pop}$ &   30.7 & 27.9  \\
per-GPU BN & 1 & $\mu_\mathcal{B}, \sigma_\mathcal{B}$ (1 img) &  \textbf{41.5} & \textbf{37.0}  \\\hline
\demph{no norm}   & \demph{2} & - &  \demph{41.1} & \demph{36.9}  \\
\demph{GN}   & \demph{2} & - &  \demph{41.4} & \demph{37.7}  \\
\end{tabular}
\vspace{.7em}
\caption{Mask R-CNN, R50-FPN with BatchNorm on heads, 3$\times$ schedule.
All entries train with a total of 16 images per iteration,
but different number of images per-GPU.
We let BatchNorm normalize the mini-batch on each GPU during training,
using a mini-batch size $N=\sum_{i=1}^{\text{img/GPU}}{R_i}$.
In testing, all models take one image at a time, and BatchNorm can use
either population statistics or mini-batch statistics (with mini-batch size $N=R_0$).
Models are evaluated by box and mask AP, averaged across 3 runs.
Results with no norm and GroupNorm (GN) are listed for reference.  }
\label{tab:maskrcnn-norm-head}
\end{table}

We experiment with a standard Mask R-CNN \cite{He2017} baseline with pre-trained ResNet-50,
implemented in \cite{wu2019detectron2}.
There is no BatchNorm (but only FrozenBN, discussed in Sec.~\ref{sec:frozenbn}) in this baseline model.
To study the behavior of BatchNorm, we replace the default \textit{2fc} box head
with a \textit{4conv1fc} head following \cite{Wu2018},
and add BatchNorm after each convolutional layer in the box head and the mask head.
The model is tuned end-to-end, while FrozenBN layers in the backbone remain fixed.
We train the model with different normalization batches, and evaluate
them with different statistics, listed in Table~\ref{tab:maskrcnn-norm-head}.

\paragraph{Mini-batch statistics outperform population statistics.}
The results in Table~\ref{tab:maskrcnn-norm-head} show that,
using mini-batch statistics at test-time obtains significantly better
results than population statistics.
In particular,
the model performs very poorly if using $\mu_{pop}, \sigma_{pop}$
in inference, but one image per GPU during training.
This is due to overfitting to patterns within the batch, and will be further
discussed in Sec.~\ref{sec:pattern-in-mini-batch}.
Also note that the sampling strategies inside R-CNN create
additional train-test inconsistency,
which may let mini-batches in inference follow a different distribution
from those in training.
This suggests room for further improvements.

To conclude, we show that using mini-batch statistics in inference
can be an effective and legitimate way to reduce train-test inconsistency.
It improves model's performance when normalization batch size is small,
and also in R-CNN's heads.

\subsection{Use Population Batch in Training}
\label{sec:frozenbn}

The previous section discusses using mini-batch statistics in inference.
As an alternative,
using population statistics in training may in theory reduce train-test inconsistency as well.
However, as \cite{Ioffe2015,Ioffe2017} observed,
normalizing by EMA during gradient descent makes the model not trainable.
One feasible approach to train with population statistics, is to use
frozen population statistics, also known as Frozen BatchNorm (FrozenBN).

\paragraph{FrozenBN} is merely a constant affine transform
$y = \frac{x-\mu}{\sqrt{\sigma^2 + \epsilon}}$, where $\mu, \sigma$ are fixed population statistics
computed ahead of time.
As a linear transform, it may lose the optimization benefits of normalization layers.
Therefore FrozenBN is often used after the model has already been
optimized with normalization, such as when
transferring a pre-trained model to downstream tasks.
Because it no longer introduces train-test inconsistency or other issues of BatchNorm,
FrozenBN is a popular alternative in areas such as object detection.

\begin{figure}[t!]\centering
  \vspace{1.em}
  \includegraphics[width=1\linewidth]{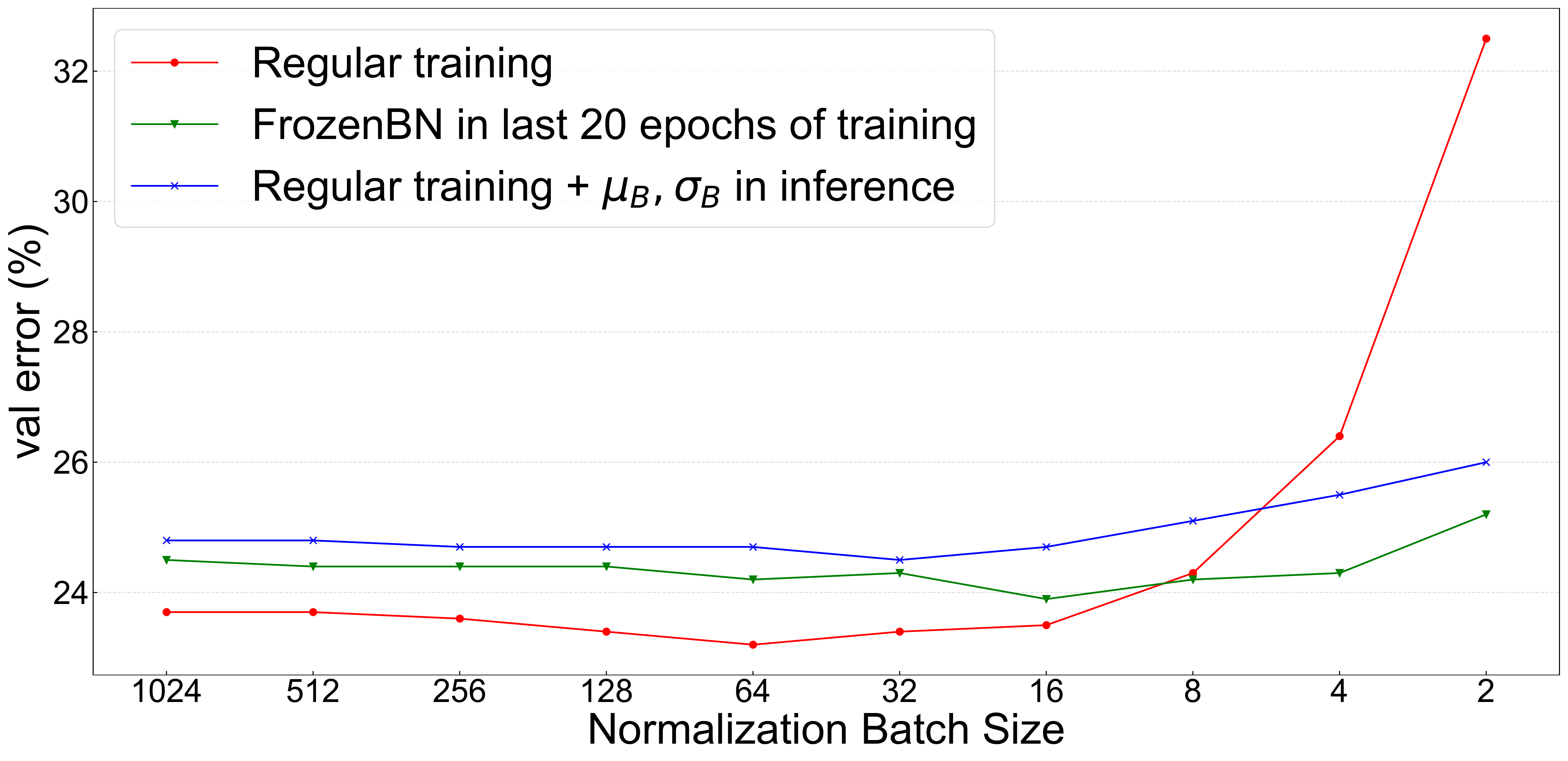}
  \caption[]{Validation error under different normalization batch sizes.
  {\color{red}Red}: train with mini-batch statistics, and inference with population statistics;
  {\color{OliveGreen}Green}: train with FrozenBN in the last 20 epochs.
  {\color{blue} Blue}: train and inference both with mini-batch statistics.
  Green and blue curves are two ways to reduce train-test inconsistency,
  which obtain similar results.}
  \vspace{-1em}
  \label{fig:nbs-frozen}
  \end{figure}

Next, we study the effect of FrozenBN in reducing train-test inconsistency.
For every model in Fig.~\ref{fig:nbs}, we take the 80th epoch checkpoint,
and train the last 20 epochs with all BatchNorm replaced by FrozenBN.
All training settings remain unchanged from the last 20 epochs of the original training,
except that we start with a linear warmup of one epoch to help optimization.
Results are shown in Fig.~\ref{fig:nbs-frozen}:
FrozenBN also effectively reduces train-test inconsistency,
getting 25.2\% error rate even when the model
is trained with a small normalization batch size of 2 for its first 80 epochs.
When normalization batch size is large enough,
tuning with FrozenBN underperforms regular BN, which is also observed in object detection domain
\cite{Peng2018,he2019rethinking}.

\subsection{Discussions and Related Works}
\label{sec:review-noise-inconsistency}
\paragraph{``Normalization batch size''} is a term we choose to avoid
confusion with other related notions of batch size.
This concept has already been implicitly used in previous works:
\cite{Goyal2017} emphasizes
that the normalization batch size should be kept unchanged despite
it studies large-batch training.
\cite{hoffer2017train} proposes to use a small normalization batch size for regularization.
\cite{wu2019multigrid,smith2017don}
use a varying total batch size, instead of a constant total batch size during
the course of training.
They both decouple normalization batch size from total batch size and adjust them separately.

We develop new understanding of
BatchNorm's behavior w.r.t. normalization batch sizes by analyzing Fig.~\ref{fig:nbs},
as it ablates three sources of performance gap:
training noise, generalization gap, and train-test inconsistency.

\paragraph{Small normalization batch size} is known to have negative effect,
as observed in previous works \cite{Ioffe2017,Wu2018}.
We analyze the underlying reasons,
and show that on ImageNet, the large performance gap between small and regular normalization batch size
can be reduced by 3.5\% with PreciseBN (Table.~\ref{tab:precisebn-batchsize}),
then by another 6.5\% by using mini-batch statistics (Fig.~\ref{fig:nbs}).
These improvements do not affect the SGD training at all.

We show that train-test inconsistency can be greatly reduced by using
mini-batch statistics in inference, or by using FrozenBN in training.
Both approaches simply change the ``batch'' to normalize, and
significantly improve performance of standard BatchNorm in
the regime of small normalization batch sizes.

Our analysis stops at a normalization batch size of 2.
The relation to InstanceNorm~\cite{Ulyanov2016} will be discussed separately in Appendix~\ref{sec:instancenorm}.

\paragraph{Mini-batch statistics in inference} can help reduce
train-test inconsistency.
To our knowledge, applying this technique in R-CNN's head has not
appeared in previous works,
but \cite{Wu2018} also observes that population statistics perform poorly
in R-CNN's head.

EvalNorm~\cite{singh2019evalnorm} is closely related to using mini-batch statistics in inference:
it aims to approximate mini-batch statistics
based on statistics of the single input instance, together with population statistics.
It reduces train-test inconsistency by a similar mechanism,
without having to access a mini-batch in inference.

Batch statistics in inference has other potential use: \cite{Isola2017}
uses batch statistics as a source of random noise.
\cite{nichol2018first} evaluates using batch statistics in inference for meta-learning.
In Sec.~\ref{sec:domain-discussions}, we review literatures that use batch statistics in inference
to address domain shift.

\paragraph{FrozenBN} exists in previous works that fine-tune a pre-trained classifier for downstream tasks
such as object detection \cite{Ren2015,He2017,Li_2019_CVPR}, metrics learning \cite{musgrave2020metric}.
When applied in fine-tuning, it's common to
freeze both normalization statistics and the subsequent affine transform,
so that they can be fused into a single affine transform
 \footnote{ However, during fine-tuning it is not a good idea to fuse the affine transform
 with other adjacent layers (e.g. convolution), although this is a standard
 optimization for deployment. See Appendix~\ref{sec:wrong-fusion}.}.
 This variant of FrozenBN is computationally efficient and often performs similarly well.

Even without the context of fine-tuning,
we have shown that using FrozenBN during training is generally helpful in
reducing train-test inconsistency,
which partly explains why it's useful in transfer learning.
The benefit of switching to FrozenBN in the middle of training has also been observed in
previous works such as \cite{johnson2018image,xie2019intriguing,krishnamoorthi2018quantizing}.

\paragraph{Reduce training noise in BN.}
Apart from reducing train-test inconsistencies, several other normalization methods
attempt to reduce training noise of BatchNorm.
BatchRenorm (BRN) \cite{Ioffe2017} introduces correction terms
to bring the training-time statistics closer to EMA,
to reduce the noise in batch statistics.
Moving Average BatchNorm (MABN) \cite{yan2020towards}
reduces training noise by applying EMA to both the batch statistics and
the back-propagated gradients.
Both methods can improve BatchNorm under the small batch regime.
\cite{shen2020powernorm,yao2020cross}
also study the noise in batch statistics and propose alternative
normalization scheme in training.

Batch-independent normalization methods, such as GroupNorm~\cite{Wu2018},
Filter Response Normalization \cite{singh2019filter}, EvoNorm~\cite{liu2020evolving}
do not suffer from training noise and have competitive accuracy.
However, unlike BatchNorm, they all incur extra cost of normalization in inference.

\paragraph{Increase training noise in BN.}
Sometimes, additional noise is desirable due to its extra
regularization effect.
Several methods that inject noise on batch statistics have been developed.
For example,
TensorFlow's \cite{Abadi2016}
BatchNorm layer has an \texttt{adjustment=} option to explicitly inject
random noise.
\cite{li2020feature} proposes to mix statistics of different samples
in order to increase noise.
\cite{Isola2017} uses mini-batch statistics at test-time
to increase noise.

%File:

\section{Batch from Different Domains}
\label{sec:mini-batch-domain}

The training process of BatchNorm models can be considered as two separate phases:
features are first learned by SGD, then the population statistics
are trained using these features, by EMA or PreciseBN. We refer to the two phases as ``SGD training'',
and ``population statistics training''.

Discrepancy between training and test dataset hurts generalization of machine
learning models.
BatchNorm models behave uniquely under such domain shift,
due to the extra phase in BatchNorm that trains
population statistics.
When data comes from multiple domains, domain gap between the
inputs in (1)
SGD training; (2) population statistics training; (3) testing can all
affect generalization.

In this section, we analyze two scenarios where domain gap appears:
when a model is trained on one domain but tested on others,
and when a model is trained on multiple domains.
We show that both can complicate the use of BatchNorm by choices of
what ``batch'' to normalize.

\subsection{Domain to Compute Population Statistics}
\label{sec:inverted-imagenet}

Domain shift occurs when the model's
training and testing phases see different data distributions.
In BatchNorm models, the extra training phase that computes population statistics
is also affected by domain shift.
Typical use of BatchNorm computes $\mu_{pop}, \sigma^2_{pop}$ on
training domain of data, but \cite{li2016revisiting} proposes ``Adaptive BatchNorm''
to compute statistics on the test domain.
Here we revisit this approach.

\paragraph{Experimental settings.}
We use a regular ResNet-50 model trained on ImageNet, following the recipe of
\cite{Goyal2017}, with a total batch size of 1024 and
a normalization batch size of 32.
To study domain shift using this model,
we evaluate it on ImageNet-C \cite{hendrycks2019benchmarking} which contains
various corrupted versions of ImageNet images.
Specifically, we use the subset of ImageNet-C with 3 different corruption types
(contrast, gaussian noise, jpeg compression) and
medium severity.
We evaluate this model's accuracy on different evaluation data,
while changing the data used to train population statistics.

\begin{table}
\newcommand{\demph}[1]{\textcolor{Gray}{#1}}
\tablestyle{1.5pt}{1.05}
\begin{tabular}{c|c|c}
Evaluation data                & Pop. stats training data & Error Rate (\%) \\ \shline
\multirow{2}{*}{IN-C-contrast} & IN                       & 40.8 \\
                               & IN-C-contrast            & \textbf{33.1} \\\hline
\multirow{2}{*}{IN-C-gaussian} & IN                       & 64.4 \\
                               & IN-C-gaussian            & \textbf{52.5} \\\hline
\multirow{2}{*}{IN-C-jpeg}     & IN                       & 40.4 \\
                               & IN-C-jpeg               & \textbf{36.7} \\\shline
\multirow{3}{*}{IN}            & IN                       & \textbf{23.4} \\
                               & IN-C-contrast            & 39.7 \\
                               & IN (val set w/o augs)    & 23.8 \\\shline
\end{tabular}
\vspace{.7em}
\caption{Classification error of one ResNet-50 model trained on ImageNet (IN).
The model is evaluated on IN and different corruption types
of ImageNet-C (IN-C), after re-training its population
statistics on different data.
As ImageNet-C does not contain training data,
we use 1000 images from each corruption type to train population statistics
and evaluate on the rest.
}
\label{tab:domain-shift}
\end{table}

\paragraph{Results.} Table~\ref{tab:domain-shift}
shows that when significant domain shift exists, the model obtains the best error
rate after training population statistics on the domain
used in evaluation, rather than the dataset used in SGD training.
Intuitively, recomputing these statistics reduces train-test inconsistency and
improves generalization on the new data distribution.

However, we also note that recomputing population statistics on target distribution
may end up hurting performance as well.
As an example, we recompute $\mu_{pop}, \sigma^2_{pop}$ of the above model on
the exact inputs to be evaluated, i.e.
the ImageNet validation set with inference-time preprocessing instead of training augmentations.
Despite the same data is observed during
population statistics training and evaluation, performance
becomes slightly worse (last row in Table~\ref{tab:domain-shift}).
We hypothesize this is due to the inconsistency among the two training steps,
Whether it is beneficial to adapt population statistics on new domain is still
a question that needs to be evaluated in different contexts.

\subsection{BatchNorm in Multi-Domain Training}
\label{sec:layer-sharing}
BatchNorm has a unique property that its outputs depend on not only individual samples,
but also how samples are grouped into batches in training. Formally,
for a BatchNorm layer $f$,
\begin{equation}\label{eqn:bn-split}
 f([X_1, X_2, \cdots X_n]) \ne [f(X_1), f(X_2), \cdots f(X_n)]
\end{equation}
where $X_1, X_2, \cdots $ are multiple input batches,
and $[\cdot, \cdot]$ means concatenation on the batch dimension.
When batches come from different domains (e.g. different datasets or different features),
the two sides of Eqn.~\ref{eqn:bn-split} determine whether to normalize different domains together,
or to normalize each domain individually. Such differences can have a significant impact.

%\begin{figure}[t]\centering
  %\vspace{1.em}
  %\includegraphics[width=0.8\linewidth]{fig/DomainSpecific.pdf}
  %\caption[]{Demonstration of how shared layers may be applied to inputs
  %from multiple domains during training. When the shared layers contain BatchNorm,
  %the two approach are not equivalent.}
  %\vspace{-1em}
  %\label{fig:domain-specific}
  %\end{figure}

\begin{figure}[t!]\centering
  \vspace{1.em}
  \includegraphics[width=1.0\linewidth]{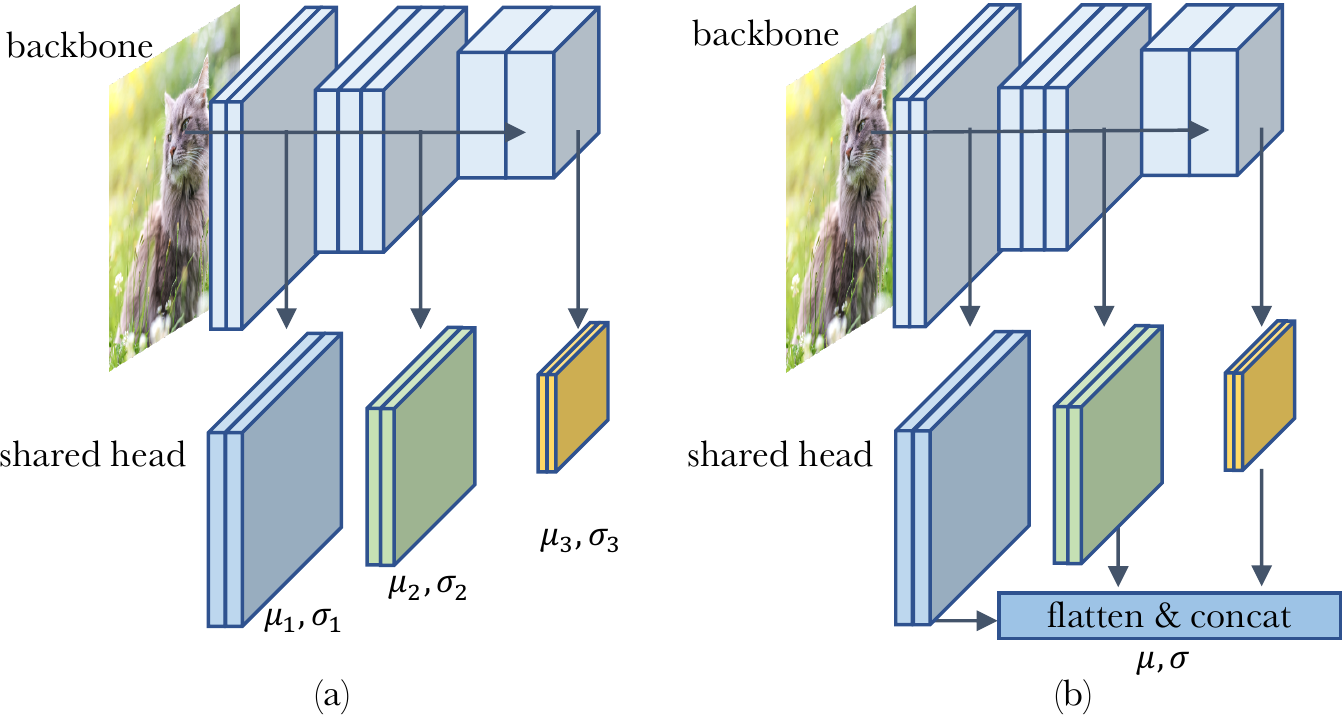}
  \caption[]{Illustration of two ways to use BatchNorm in RetinaNet's shared head,
  which is applied on 5 features from backbone (only 3 are shown).
  (a) compute different statistics for each input feature;
  (b) compute one set of statistics using all features combined.
}
  \label{fig:retinanet}
\end{figure}

\paragraph{BN in RetinaNet's Head.} We perform a case study using the RetinaNet \cite{Lin2017a} object detector,
equipped with Feature Pyramid Networks (FPN) \cite{Lin2017}.
This model has a sequence of convolutional layers (a.k.a. ``head'')
that makes predictions
on 5 features from different feature pyramid levels,
illustrated in Fig.~\ref{fig:retinanet}.
The head is shared across all 5 feature levels,
which means it is trained with inputs from 5 different distributions or domains.
We add BatchNorm and channel-wise affine layer after all intermediate
convolutional layers in this shared head, and study its behavior.

In this model, the input features of the head $X_1 \cdots X_5$ come from different
FPN levels, and have different spatial dimensions.
Therefore,
it is straightforward to train the model in the form of Fig.~\ref{fig:retinanet}(a)
where the head is applied to features independently, each normalized by its own statistics.
However, Fig.~\ref{fig:retinanet}(b) is a valid choice:
we flatten and concatenate all input features, in order to compute a single
set of statistics that normalize all features together.
We refer to these two choices as using
``domain-specific statistics'' and ``shared statistics'' in SGD training.

Similar choices exist in how to train population statistics.
We can compute unique population statistics for each feature level,
and use them in inference based on the source of inputs, similar to Fig.~\ref{fig:retinanet}(a).
Or we can normalize all feature levels with the same set of population statistics.

%To reduce train-test inconsistency, we argue that SGD training and
%population statistics training shall make consistent choices.
%To verify our claim,
We explore the combination of the above choices
in this shared head, to show how inconsistent choices in the two training phases
affect the model's generalization.
In addition, we also experiment with the choice to share the affine transform layers
after BatchNorm, or to use different affine transform for each input feature.

\begin{table}
\newcommand{\demph}[1]{\textcolor{Gray}{#1}}
\tablestyle{1.5pt}{1.05}
\begin{tabular}{c|c|c|c|c}
Norm in Head & SGD Training & Population Stats & Affine Parameters & AP \\ \shline

SyncBN & shared & shared & shared          & \textbf{39.1} \\
SyncBN & shared & domain-specific & shared          & 2.1 \\\hline
SyncBN & domain-specific & shared & shared          & 34.1 \\
SyncBN & domain-specific & domain-specific & shared & \textbf{39.1} \\\hline
SyncBN & domain-specific & shared & domain-specific & 35.5 \\
SyncBN & domain-specific & domain-specific & domain-specific & \textbf{39.2} \\\shline
\demph{GN} & \demph{-} & \demph{-} & \demph{shared} & \demph{39.5} \\
\demph{no norm} & \demph{-} & \demph{-} & \demph{-} & \demph{38.7} \\
\end{tabular}
\vspace{.3em}
\caption{Object detection performance (AP) of RetinaNet,
using R50-FPN, 3$\times$ schedule implemented in \cite{wu2019detectron2}.
All BatchNorm entries are trained with a mini-batch size of 16 images and SyncBN.
The model works well only when SGD training and population statistics training follow
consistent paradigm (row 1, 4, 6).
Standard usage of BatchNorm (row 3) leads to drastic performance drop.
Popular baselines have no normalization in RetinaNet heads due to
difficulty to use BatchNorm correctly.
Results with GroupNorm (GN) and no normalization are listed for reference.
}
\label{tab:retinanet-norm-head}
\vspace{-1em}
\end{table}

\paragraph{Consistency is crucial.}Table~\ref{tab:retinanet-norm-head} lists
experimental results of these combinations.
In summary, we show that for a BatchNorm that's used multiple times on
different domains / features,
the way its population statistics are computed should be consistent with
how features are normalized during SGD optimization.
Otherwise, such gap causes the features to not generalize
in inference.
Meanwhile, whether to share the affine layer or not has little impact.

We emphasize that a standard implementation of this head
would perform poorly.
Row 3 in Table~\ref{tab:retinanet-norm-head} corresponds to a straightforward implementation,
which would: (1)
apply the shared head separately on each feature,
because this is how other (convolutional) layers in the head are applied;
(2) maintain only one set of population statistics because the
layer is conceptually ``shared'' across features.
Appendix~\ref{sec:impl-retinanet} uses pseudo-code to further demonstrate
that the correct usages of BatchNorm are less straightforward to implement,
and therefore less common.

Though we use RetinaNet with shared layers as the example, we note that the issue
we discuss also exists when training a model on multiple datasets:
the choice between Fig.~\ref{fig:retinanet}(a) and (b) remains.
Compared to directly training on multiple datasets,
the multi-domain nature of shared layers can be easily overlooked.

\subsection{Discussions and Related Works}
\label{sec:domain-discussions}

\paragraph{Domain Shift.} Sec.~\ref{sec:inverted-imagenet} briefly discusses the role of population statistics
in domain shift.
The idea of recomputing population statistics on target domain
is first proposed as ``Adaptive BatchNorm'' \cite{li2016revisiting}.
This is followed by other works that modify BatchNorm for domain adaptation
such as \cite{cariucci2017autodial}.
Defense against image corruptions or adversarial examples can be seen as
special cases of domain adaptation,
and \cite{benz2021revisiting,schneider2020improving} show that
adaptive population statistics help in these tasks as well.
Moreover, when test-time inputs are given as a sufficiently large mini-batch,
the mini-batch itself can be used as the ``population'' of target domain.
\cite{nado2020evaluating, awais2020towards} show that
recomputing statistics directly using the test-time mini-batch
also improves robustness against image corruptions or adversarial attacks.

All above methods require access to sufficient amount of target domain data,
which is often unavailable.
Therefore, models with BatchNorm can suffer from domain shift more than
other normalization methods.
\cite{alex2019big} and \cite{pan2018two} both observe that
the generalization gap of BatchNorm models in transfer learning
can be improved after incorporating batch-independent normalization.

\textbf{Domain-specific training} of BatchNorm
on a mixture of multi-domain data has been proposed frequently
in previous works, under the name of
``Domain-Specific BN'' \cite{chang2019domain},
``Split BN'' \cite{zajkac2019split},
``Mixture BN'' \cite{xie2019intriguing},
``Auxiliary BN'' \cite{xie2019adversarial},
``Transferable Norm''\cite{wang2019transferable}.
Domain-specific population statistics in RetinaNet
has been used in \cite{tensorflow_od_api}.
These methods all contain some of the following three choices.
\begin{itemize}[leftmargin=*]
  \item \textbf{Domain-specific SGD training}:
    whether $\mu_\mathcal{B}, \sigma_\mathcal{B}$ used in gradient descent training are computed from one
    or multiple domains.
  \item \textbf{Domain-specific population statistics}:
    whether to use one set of $\mu_{pop}, \sigma_{pop}$
    in testing for all domains, or one for each domain.
    This requires knowing the domain of the inputs at test-time.
    Such knowledge is trivial if the multiple domains come from layer sharing
    (Sec.~\ref{sec:layer-sharing}),
    but not necessarily available in other applications, e.g. testing on
    inputs from multiple datasets.
  \item \textbf{Domain-specific affine transform}:
    whether to learn domain-specific affine transform parameters,
    or use a shared affine transform.
    This also requires knowing the domain of inputs at test-time.
\end{itemize}

By ablating the above three choices, we show that it is important to use
consistent strategy between SGD training and population statistics training,
although such implementation may appear unintuitive.

Machine learning models that deal with multiple domains of data
are common. Some algorithms are even specifically designed to handle
such data, e.g. shared layers, adversarial defense, GANs, etc.
Our analysis in this section shows the importance to carefully
consider what domain of data to normalize.

%\item \textbf{GANs} \cite{Goodfellow2014} train a discriminator to distinguish true samples and fake samples.
%The discriminator is therefore trained with data from two distributions.
%\cite{Chintala2016} proposes to avoid mixing two distributions in a batch when training GANs.
%\wyx{TODO No evidence for the claim. Papers today don't use BN in GANs.}

%File:

\section{Information Leakage within a Batch}
\label{sec:info-leak}

We describe another type of BatchNorm's caveats as ``information leakage'',
when a model learns to utilize information it is not designed to use.
Information leakage happens because BatchNorm does not make independent
predictions for every input sample in a mini-batch. Instead, prediction
of each sample makes use of other samples through the mini-batch statistics.

When other samples in a mini-batch carry useful information,
models may learn to exploit such mini-batch information instead of learning
representations that generalize for individual samples.
Such behavior may appear in various applications.
We review a few examples in this section.

\subsection{Exploit Patterns in Mini-batches}
\label{sec:pattern-in-mini-batch}

\begin{figure}[t]\centering
  \includegraphics[width=0.95\linewidth]{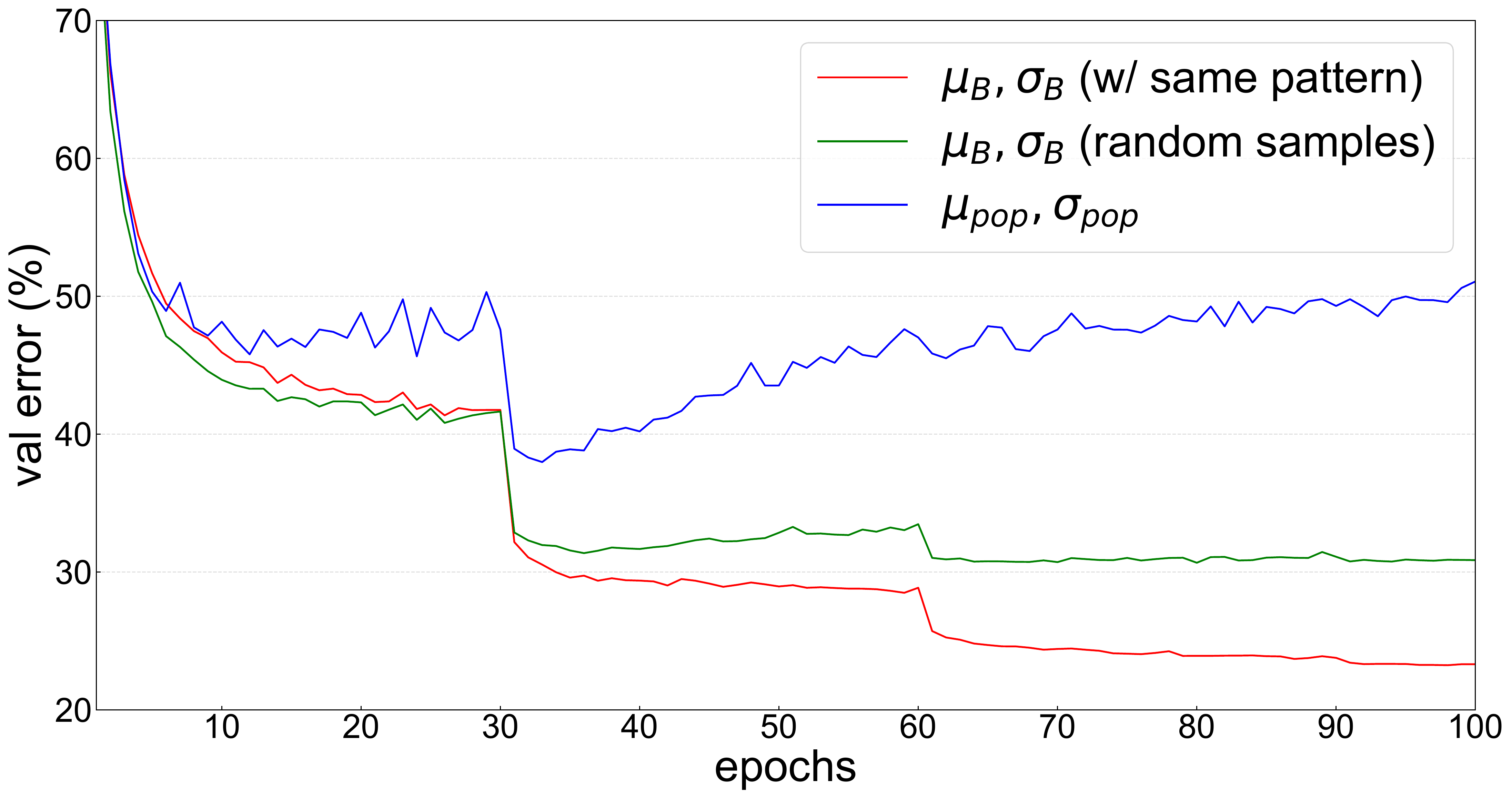}
  \caption{Validation curves of a model trained with a special pattern in the construction of mini-batches.
  When evaluation uses $\mu_\mathcal{B}, \sigma_\mathcal{B}$ with the same mini-batch patterns ({\color{red} red}),
  the model makes good predictions.
  When evaluation uses
  $\mu_\mathcal{B}, \sigma_\mathcal{B}$ with randomly sampled mini-batches ({\color{OliveGreen}green}),
  or
  uses $\mu_{pop}, \sigma_{pop}$ ({\color{blue}blue}, PreciseBN),
  results show sign of severe overfitting. }
  \label{fig:baddist}
  \vspace{-1em}
  \end{figure}

\paragraph{Artificial pattern in mini-batch.} \cite{Ioffe2017} designs an
experiment,
where an ImageNet classifier is trained with a normalization batch size of 32,
but every such batch is hand-crafted with 16 classes and 2 images per class.
Clearly, when the model makes predictions for a sample during training,
it can lower the training loss by utilizing the fact
that another sample in the mini-batch must have the same label.
As a result, the model exploits this pattern, and does not generalize after training.
Fig.~\ref{fig:baddist} shows results of such an experiment performed on ResNet-50,
with all other experimental details following \cite{Goyal2017}.

In this example, the problem can be addressed by adopting
"Ghost BatchNorm" (Appendix~\ref{sec:changing-nbs}) during training
to normalize over a subset of 16 images from different classes,
so that each normalization batch no longer contains special patterns.
Next, we look at a realistic and common scenario where this issue appears,
and show other practical remedies.

\begin{figure}[t]\centering
  \includegraphics[width=0.9\linewidth]{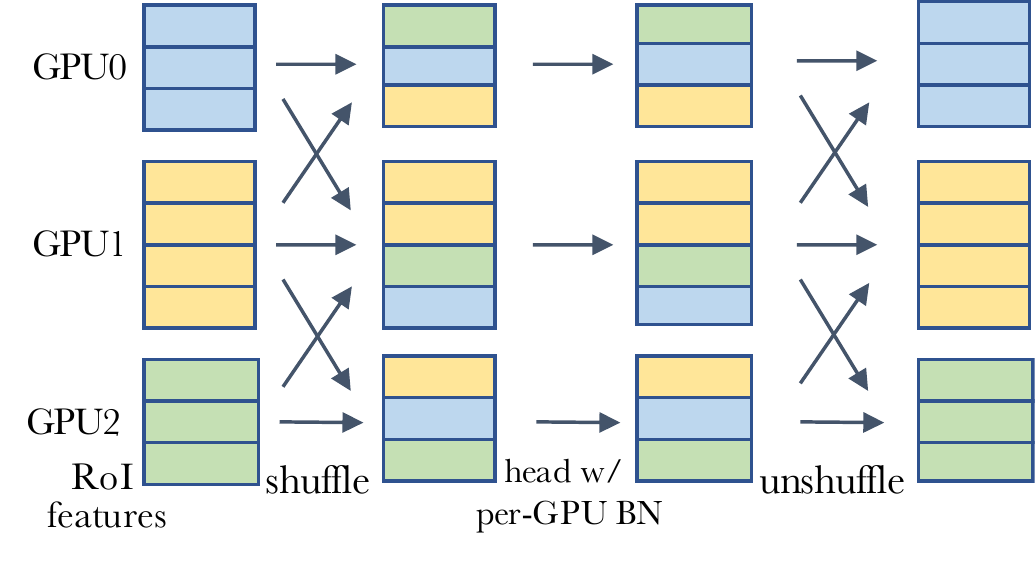}
  \caption[]{Illustration of computation of an R-CNN's head with shuffled inputs.
  Each GPU is initially given one image and all its RoIs, filled with the same color.
  RoI features are first shuffled among GPUs, to prevent the per-GPU BatchNorm in the head
  from exploiting information between RoIs.
  Other computation in the head is independent for each RoI.
  After the head, all RoI predictions are shuffled back to
  form predictions for each image.
  }
  \label{fig:shuffle-rcnn}
\end{figure}

\paragraph{Correlated samples in R-CNN's head.} We study the second stage (head) of a Mask R-CNN model, under the same
setting as Sec.~\ref{sec:maskrcnn-norm-head}.
We have shown in Sec.~\ref{sec:maskrcnn-norm-head} that adding BatchNorm in the
head leads to significantly degraded performance, if population statistics are
used in inference.
We claim that the cause is similar to Fig.~\ref{fig:baddist}:
when doing per-GPU BatchNorm in the head, samples in a mini-batch are formed
by RoIs from the same image.
These RoIs carry useful representations about each other that can leak through
mini-batch statistics.
Sec.~\ref{sec:maskrcnn-norm-head} uses mini-batch statistics in inference,
which allows the model to take advantage of the leakage in inference,
hence improves performance.
Here, we aim to avoid the information leakage instead.

Although the mini-batches on each GPU
contain highly correlated
RoIs in the head, RoIs between GPUs are independent.
One solution is to use SyncBN (Appendix~\ref{sec:changing-nbs})
to normalize over all RoIs across all GPUs.
By combining samples from other images, the correlation between samples within
a normalization batch is weakened.

Another solution is to randomly shuffle the RoI features among
GPUs before entering the head.
This assigns each GPU a random subset of samples to normalize,
and also weakens the correlation between samples within a normalization batch.
This process is illustrated in Fig.~\ref{fig:shuffle-rcnn}.

Experimental results in Table~\ref{tab:maskrcnn-norm-shuffle}
demonstrate that both shuffling and SyncBN effectively
address information leakage, allowing the head
to generalize well at test-time with population statistics.
In terms of speed, we note that shuffling requires fewer cross-GPU synchronization for a deep model,
but each synchronization transfers more data than what a SyncBN layer does. Therefore their
relative efficiency varies depending on the model architecture.

\begin{table}
\newcommand{\demph}[1]{\textcolor{Gray}{#1}}
\tablestyle{1.5pt}{1.05}
\begin{tabular}{c|c|c|cc}
Norm on Heads & \begin{tabular}{c}img / GPU \\ (training) \end{tabular}  & Stats. in inference &
\apbbox{~} &
\apmask{~} \\
\shline

per-GPU BN & 1 & $\mu_{pop}, \sigma_{pop}$ &   30.7 & 27.9  \\
per-GPU BN & 2 & $\mu_{pop},\sigma_{pop}$ &   40.1 & 36.2  \\
per-GPU BN & 2 & $\mu_{B}, \sigma_{B}$ &   41.3 & 37.0  \\\hline
per-GPU BN w/ shuffle & 1 & $\mu_{pop},\sigma_{pop}$ & \textbf{41.7} & \textbf{37.7} \\
per-GPU BN w/ shuffle & 2 & $\mu_{pop},\sigma_{pop}$ & \textbf{41.6} & \textbf{37.6} \\\hline
SyncBN & - & $\mu_{pop}, \sigma_{pop}$      & \textbf{41.7} & \textbf{37.7} \\
\demph{no norm} & \demph{2} & -      & \demph{41.1} & \demph{36.9} \\

\end{tabular}
\vspace{.7em}
\caption{Mask R-CNN, R50-FPN with variants of BatchNorm on heads, 3$\times$ schedule.
All other settings follow Table.~\ref{tab:maskrcnn-norm-head}.
Unlike Table.~\ref{tab:maskrcnn-norm-head},
inference using $\mu_{pop}, \sigma_{pop}$ is now able to obtain good performance,
after training with shuffled inputs or SyncBN.
The effectiveness of SyncBN also explains why row 2 is significantly better than row 1.
Results with no norm are listed
for reference: appropriate use of BatchNorm improves upon
the common baseline that uses no norm.
}
\label{tab:maskrcnn-norm-shuffle}
\end{table}

Patterns in mini-batches may appear in other applications as well. For example,
in reinforcement learning,
training samples obtained by one agent in the same episode,
if put into one mini-batch, may also exhibit correlations that can be exploited.
In video understanding models,
a mini-batch may contain multiple frames cropped from the same video clip.
If such applications are affected by the use of BatchNorm on correlated mini-batches,
they may also benefit from a careful implementation that
ensures normalization batches don't contain correlated samples.

\subsection{Cheating in Contrastive Learning}
In contrastive learning or metric learning,
the training objective is often designed to compare samples within a mini-batch.
Samples then simultaneously play the role of both inputs and labels in the loss function.
In such cases, BatchNorm can leak information about other samples,
and hence allow the model to cheat during training, rather than to learn
effective representations.

This cheating behavior is observed in \cite{Henaff2019},
which adopts LayerNorm~\cite{Ba2016} instead.
Other works in contrastive learning address the issue by using BatchNorm with
different implementations, similar to what we study in Sec.~\ref{sec:pattern-in-mini-batch}.
MoCo~\cite{he2019momentum}
shuffles the samples among mini-batches,
to prevent BatchNorm from exploiting other samples.
SimCLR~\cite{chen2020simple}, BYOL~\cite{grill2020bootstrap}
use SyncBN to address this issue.

Other remedies exist thanks to the property of contrastive learning.
In many teacher-student contrastive learning frameworks,
one branch of the model does not propagate gradients.
Taking advantage of this property,
\cite{li2021momentum,cai2021exponential} propose to prevent leakage by
not using mini-batch statistics, but using
moving average statistics in this branch. This also effectively avoids
cheating.

%File:

\section{Conclusions}
\label{sec:discussions}

This paper discusses a number of practical issues when BatchNorm is applied in
visual recognition tasks.
Here we summarize them under different perspectives.

\paragraph{Inconsistency.}
Train-test inconsistency plays a crucial role in BatchNorm's behavior.
In Sec.~\ref{sec:population-statistics}
we discuss ways to compute population statistics that are more accurate and
do not lag behind train-time mini-batch statistics.
In Sec.~\ref{sec:train-test-inconsistency}, we show that train-test inconsistency
can be reduced by using mini-batch statistics in inference or using
population statistics in training.
In Sec.~\ref{sec:mini-batch-domain},
we show the importance of consistency between what batches
to perform optimization, what data to compute population statistics,
and what data to evaluate models.
The information leakage issue in Sec.~\ref{sec:info-leak} arises from
the incosistency between training with mini-batches and
testing with individual samples.
Moreover, train-test inconsistency from other sources may
also interact with BatchNorm, as discussed in Sec.~\ref{sec:population-discussions}.

\paragraph{Concept of ``batch''.}
We have reviewed various unconventional usages of BatchNorm that can address its caveats.
They all simply change how the batch is constructed under different contexts.
In Sec.~\ref{sec:population-statistics},
we define the whole population as one batch,
and studies methods to compute its statistics.
In Sec.~\ref{sec:nbs},
we show that normalization batch can be different from the mini-batch used
in gradient descent.
Sec.~\ref{sec:batch-stats-in-inference},~\ref{sec:frozenbn} show that
the batch can be freely chosen as either
a mini-batch or the population, in both training and testing.
Sec.~\ref{sec:inverted-imagenet} discusses the data used to construct the batch.
Sec.~\ref{sec:layer-sharing} focuses on the distinction between
a per-domain mini-batch and a joint mini-batch across all domains.
Sec.~\ref{sec:info-leak} shows that when the nature definition of batch has undesired
properties, we can change the batch by techniques like shuffling.

Unlike other alternative normalization schemes proposed before,
all above approaches still normalize on the batch dimension of features,
but with different ways to construct batches. We aim to illustrate the importance
to carefully consider the concept of \emph{batch} in every scenario.

\paragraph{Library implementations.}
Many libraries have provided ``standard'' implementations of BatchNorm.
However, as we see in this paper, using an off-the-shelf implementation
may be suboptimal, and alternative ways to use BatchNorm may not be readily available
in such standard implementations.
For example, they use EMA
and often do not provide ways to re-estimate population statistics.
This causes difficulties in using precise statistics (Sec.~\ref{sec:precisebn})
or in domain shift (Sec.~\ref{sec:inverted-imagenet}).
Standard implementations maintain one set of population statistics,
but per-domain statistics can be useful according to Sec.~\ref{sec:mini-batch-domain}.
Standard implementations use per-GPU mini-batch statistics in training,
and population statistics in testing, but as Sec.~\ref{sec:train-test-inconsistency}
shows, these are not necessarily the best choices.
Issues in Sec.~\ref{sec:info-leak} are addressed by implementation techniques
such as shuffling and SyncBN.

Moreover, due to the unique properties of BatchNorm,
it often introduces non-standard interactions with the underlying deep learning
system, and causes other practical issues or bugs.
Appendix~\ref{sec:interaction-with-libraries} lists a few examples.
We hope our review raises awareness of the limitations and potential
misuse of a standard BatchNorm implementation.

\paragraph{Practical implications.} Many of the caveats we discussed in this paper
may appear uncommon in practice.
And in fact, a regular form of BatchNorm does perform remarkably well when applied
in the following common scenario: supervised training with proper batch size and
training length, using i.i.d.~fixed-size batches randomly drawn from a single dataset,
trained without layer sharing, and tested on data of similar distribution
after the model has converged.

However, as BatchNorm and CNNs are widely used in other applications
and learning paradigms,
the caveats of BatchNorm deserve attention.
This review discusses these caveats in the limited context of visual representation learning,
using standard vision models in image classification and object detection as testbed.
When applied in other tasks, BatchNorm's unique properties may present new challenges
that need to be further investigated.
We hope this review helps researchers reason carefully about their use of
BatchNorm, to not hide or hinder new research breakthrough.

\section*{Acknowledgements}
We would like to thank Kaiming He, Ross Girshick, Piotr Doll{\'a}r, Andrew Brock,
Bjarke Roune for useful discussions and feedbacks.

\newpage
\appendix
%File:

\section{Appendix}

\subsection{PreciseBN and EMA with $\lambda=0$}
\label{sec:large-batch-ema-lambda}
Given that PreciseBN works with $N \le 10^4$ samples,
one may assume that EMA can work similarly to PreciseBN under large-batch training,
as long as momentum $\lambda$ is set to 0 to discard all historical statistics.
We note that this is often not true.

First, even when SGD training uses a large
mini-batch size, normalization typically still uses smaller batch
either to balance regularization (see more in Sec.~\ref{sec:nbs}) or
due to large communication cost across workers.
Second, in typical training implementations,
an EMA update is always followed by an SGD optimization step, which
updates the representations and leaves the population statistics outdated again.
This is unlike PreciseBN, which computes population statistics
\textit{after} SGD optimization.

Therefore, although EMA with $\lambda=0$ is conceptually equivalent to PreciseBN,
it is only comparable to PreciseBN under large-batch training,
and requires overcoming its own implementation challenges.

\subsection{Estimators of Population Variance}
\label{sec:variance}
Given $N=k\times B$ i.i.d.~samples $x_{ij} \in \mathbb{R}, i=1\cdots k, j=1\cdots B$ drawn from a random variable $X$, grouped into $k$ batches of batch size $B$,
we'd like to estimate the variance of $X$ using per-batch mean $\mu_i$ and per-batch variance $\sigma_i^2$, defined as:
\[ \mu_{i} = \sum_{j=1}^B \frac{x_{ij}}{B}, \sigma_i^2=\sum_{j=1}^B\frac{(x_{ij} - \mu_i)^2}{B} \]
One estimator we can use is:
\begin{align}\label{eqn:var-ours}
\hat{\sigma}^2 &= \frac{N}{N-1}\left[ \sum_{i=1}^k \frac{\mu_i^2 + \sigma_i^2}{k} - (\sum_{i=1}^k\frac{\mu_i}{k})^2  \right]
\end{align}
The estimator originally used in \cite{Ioffe2015} is:
\begin{equation}\label{eqn:var-original}
\hat{\sigma}^2 = \frac{B}{B-1}\sum_{i=1}^k \frac{\sigma_i^2}{k}
\end{equation}

Both estimators are unbiased, but they have different variances. By substituting the ``variance of sample variance'' \cite{sample_variance},
the variance of Eqn.~\ref{eqn:var-original} is given by:
\begin{align*}
 \mathrm{Var}[\hat{\sigma}^2] &=\frac{1}{k^2}\sum_{i=1}^k\mathrm{Var}[\frac{B}{B-1}\sigma_i^2]
= \frac{1}{k}\mathrm{Var}[\frac{B}{B-1}\sigma_0^2] \\
& = \frac{1}{k}\frac{\sigma^4}{B}(\kappa-1+\frac{2}{B-1}) = \frac{\sigma^4}{N}(\kappa - 1 + \frac{2}{B-1})
\end{align*}
where $\sigma, \kappa$ are standard deviation and kurtosis of $X$.
Therefore, a small $B$ increases variance of the estimator in Eqn.~\ref{eqn:var-original}.

Through basic algebra it can be shown that Eqn.~\ref{eqn:var-ours} computes the unbiased
sample variance of all $N$ samples, which is equivalent to Eqn.~\ref{eqn:var-original} when $k=1, B=N$.
Because $B<<N$ in practice, this proves that Eqn.~\ref{eqn:var-ours} is a better estimator.

When Eqn.~\ref{eqn:var-ours} is applied in a BatchNorm layer using a population of $N$ input features of shape $(H, W)$, the total number of samples
becomes $N\times H\times W$, which is large enough such that the bessel correction factor
is negligible.

\subsection{Compute Population Statistics Layer-by-Layer}
\label{sec:layer-wise-precise}
By definition, population statistics shall be computed by forwarding the entire population
as one batch. However, with a large population size $N$, this is often not practical due to memory constraints.

On the other hand, splitting to batches of size $B$ and
running $\frac{N}{B}$ forward passes will result in different outputs.
Although the population statistics of the first BatchNorm layer can be accurately
computed by aggregating moments of each batch (Eqn.~\ref{eqn:var-ours}),
deeper layers are affected, as mentioned
in Sec.~\ref{sec:precisebn}.
We need a way to compute true population statistics so we can tell whether our
approximation by split and aggregation is sufficient.

To accurately compute population statistics with limited memory,
we can estimate the statistics layer-by-layer.
To compute the statistics of the $k$th BatchNorm layer, we first
obtain true population statistics for its previous $k-1$ BatchNorm layers.
Then, by letting the previous $k-1$ layers use population statistics
and the $k$th layer use batch statistics,
forwarding the entire population at any batch size
produce batch statistics of the $k$th layer that can be later aggregated
into population statistics by Eqn.~\ref{eqn:var-ours}.
This algorithm allows us to use any batch size
to compute true population statistics, as if the population is one batch.

Such layer-wise algorithm has a time complexity that's quadratic w.r.t. the depth of model,
and is therefore quite expensive.
On the other hand, forwarding the whole model $\frac{N}{B}$ times with batch size $B$ is cheaper.
As shown in Table~\ref{tab:precisebn-batchsize}, with a large enough $B$, the later approximation is good enough.

\subsection{PreciseBN in Reinforcement Learning}
\label{sec:ema-rl}

Sec.~\ref{sec:ema-failures} shows that PreciseBN can be beneficial when
we run inference during training, e.g., to draw a validation curve.
However, apart from drawing curves for monitoring, most supervised training tasks
do not require using the model in inference mode during training.

On the other hand, in reinforcement learning (RL),
most algorithms require running inference along with training, in order
to interact with the environment and collect experiences as training data.
Therefore RL algorithms may suffer from poor estimation of EMA:
as the model is evolving, EMA may fail to accurately reflect the statistics.
As an example, the appendix of ELF OpenGO \cite{tian2019opengo}
describes its use of PreciseBN in AlphaZero training.
They report noticeable improvement in results after using PreciseBN to address
the issue they referred to as ``moment staleness''.

\subsection{Changing Normalization Batch Size}
\label{sec:changing-nbs}

In a classic implementation of BatchNorm layers, the normalization batch size is equal to the
per-GPU (or per-worker) batch size.
This makes it difficult to change normalization batch size:
a large per-GPU batch size requires larger memory consumption,
and a small per-GPU batch size is typically inefficient on modern hardwares.
Therefore, a few implementations of BatchNorm have been developed to change
normalization batch size regardless of the per-GPU batch size.

\paragraph{Synchronized BatchNorm (SyncBN)}, also known as Cross-GPU BatchNorm \cite{Peng2018}
or Cross-Replica BatchNorm~\cite{brock2018large},
is an implementation of BatchNorm that uses large normalization batch size up to the
SGD batch size in data-parallel training.
It is effective in tasks where the per-GPU batch size is limited,
such as object detection \cite{Peng2018, he2019rethinking} and semantic segmentation \cite{Zhang_2018_CVPR}.
In SyncBN, workers perform collective communication in order to
compute and share the mean and variance of a larger batch that spans across
multiple workers.
We discussed some implementation details of SyncBN in Appendix~\ref{sec:syncbn-impl}.

\paragraph{Ghost BatchNorm}\cite{hoffer2017train} is an implementation of BatchNorm
that reduces normalization batch size,
simply by splitting the batch into sub-batches and normalizing them individually.
When per-GPU batch size is large, this is found to
improve regularization, as observed in \cite{cifar10in1m,smith2017don}.
Such empirical evidence matches our analysis in Sec.~\ref{sec:train-test-inconsistency}.

\label{sec:virtualbn}
\paragraph{Virtual BatchNorm}\cite{salimans2016improved}
is an expensive way to increase normalization batch sizes without synchronization
between GPUs: it simply uses extra input images to compute the batch statistics.
Note that this is different from increasing per-GPU batch size:
the extra input images does not contribute gradients, therefore does not have a
significant memory cost.
But it still requires extra forward computation cost.
Similar idea was proposed in \cite{guo2018double}.
Virtual BatchNorm can also be a legitimate (but expensive) way to use mini-batch
statistics at test-time, by including extra training images to form a
virtual batch.

\paragraph{Gradient Accumulation.}
We note that the common ``Gradient Accumulation'' technique
(historically known as ``iter\_size'' in Caffe \cite{Jia2014} ) does \emph{not}
change normalization batch size.
It can however increase SGD batch size given fixed
per-GPU batch size and fixed number of GPUs.

\subsection{Implementation of SyncBN}
\label{sec:syncbn-impl}
When data-parallel workers (or GPUs) have the same batch size, a simple implementation of
SyncBN only needs to execute one \texttt{all-reduce} operation of size $2\x \text{\#channels}$ to compute
mini-batch statistics $\mu=\mathrm{E}_\mathcal{B}[x]$ as well as $\mathrm{E}_\mathcal{B}[x^2]$,
across the large batch $\mathcal{B}$ distributed among workers.
Then the variance can be computed using $\mathrm{Var}_\mathcal{B}[x]=\mathrm{E}_\mathcal{B}[x^2]-\mathrm{E}_\mathcal{B}[x]^2$.

In certain applications, the definition of ``batch size'' is more vague.
For example, in object detection, workers often have the same
number of images, but different number of pixels per image, or
different number of RoIs (for R-CNN style models) per-image.
In these scenarios, workers may end up with different number of elements
that needs to be normalized.
Instead of the aforementioned simple implementation, SyncBN can consider the
differences in mini-batch sizes among workers.
However we note that this does not appear necessary and empirically does not lead to
significant differences in results.

We also note that an implementation of SyncBN can be error-prone and requires
careful verification.
An implementation of SyncBN was added to PyTorch in Mar. 2019,
but did not produce correct gradients
as first
\href{https://github.com/facebookresearch/detectron2/blob/989f52d67d05445ccd030d8f13d6cc53e297fb91/detectron2/layers/batch_norm.py#L150-L152}{reported}
in detectron2 and later \href{https://github.com/pytorch/pytorch/pull/36382}{fixed in Apr. 2020}.
An implementation was added to TensorFlow Keras in Feb. 2020, but it does
not produce correct outputs \href{https://github.com/tensorflow/tensorflow/commit/da8b395cf792a6585bc73b65166beebba6be13f6}{until Oct. 2020}.

\subsection{InstanceNorm and BatchNorm}
\label{sec:instancenorm}
InstanceNorm (IN) \cite{Ulyanov2016} can be roughly seen as BatchNorm with
normalization batch size 1, but with some key differences:
(1) During testing, IN computes instance statistics in a way that's consistent with training;
(2) During training, the output of each sample deterministically depends on its input,
 without noise or randomness that comes from other samples in the mini-batch.
Therefore, the inconsistency and regularization analysis
in Sec.~\ref{sec:nbs} no longer applies to IN.
As a reference, under the same training recipe as the experiments in
Sec.~\ref{sec:nbs}, IN obtains validation error of $\sim$28\% according to \cite{Wu2018}.

\subsection{Affine Fusion Before Fine-tuning}
\label{sec:wrong-fusion}
When FrozenBN is used, the weights of this constant affine layer can be fused with
adjacent linear layers, such as convolutional layers.
Such fusion is a standard technique for deployment,
because the model performs mathematically equivalent computation after the fusion.
However, we remind readers that such fusion is no longer mathematically equivalent in fine-tuning.

We use a toy example to demonstrate this:
let's minimize a function $J = (\lambda x)^2 $  w.r.t. $x$
using gradient descent with a step size of 1,
where $\lambda = 0.5$ is a frozen constant and $x$ is initialized somewhere nonzero.
Because $\frac{\partial J}{\partial x} = 2\lambda^2x = \frac{x}{2}$,
each GD step is $x \leftarrow x - \frac{x}{2}$ which will converge to $x \rightarrow 0$.
However, if the frozen $\lambda$ is fused with $x$ for optimization,
the optimization task becomes $\text{minimize}_x J = x^2$ under a new initialization of $x$.
Then, each GD step becomes $x \leftarrow -x$, which does not converge.
In reality, a linear reparameterization like fusion can dramatically change the effective step size of certain
parameters, making a model harder to optimize.
Due to this, fusing a FrozenBN with adjacent layers
can lead to detrimental results in fine-tuning.

\subsection{Implementation of RetinaNet head}
\label{sec:impl-retinanet}
In Alg.\ref{alg:retinanet-head}, we provide reference implementations of the RetinaNet head
that correspond to row 1, 3, 6 of Table~\ref{tab:retinanet-norm-head},
using pseudo-code in PyTorch style.
In order to implement the head given the BatchNorm abstractions currently
available in deep learning libraries,
\texttt{row1} introduces significant complexity
and \texttt{row6} no longer ``shares'' the same head across inputs.
It is straightforward to implement it in the style of \texttt{row3} which leads to
poor performance.

An alternative implementation of \texttt{row6}
is by a layer that cycles through multiple statistics
every time it is called.
However this then introduces an extra state of the layer that depends on how
many times it is called, which may result in other complications.

\begin{algorithm}[t]
\caption{Reference pseudocode of RetinaNet head corresponding to row 1, 3, 6 of Table~\ref{tab:retinanet-norm-head}.}
\label{alg:retinanet-head}
\lstset{
  backgroundcolor=\color{white},
  basicstyle=\fontsize{8pt}{9pt}\fontfamily{lmtt}\selectfont,
  columns=fullflexible,
  breaklines=true,
  captionpos=b,
  commentstyle=\fontsize{8pt}{9pt}\color{codegray},
  keywordstyle=\fontsize{8pt}{9pt}\color{codegreen},
  stringstyle=\fontsize{8pt}{9pt}\color{codeblue},
  %frame=tb,
  otherkeywords = {self},
}
\begin{lstlisting}[language=python]
class RetinaNetHead_Row3:
  def __init__(self, num_conv, channel):
    head = []
    for _ in range(num_conv):
      head.append(nn.Conv2d(channel, channel, 3))
      head.append(nn.BatchNorm2d(channel))
    self.head = nn.Sequential(*head)

  def forward(self, inputs: List[Tensor]):
      return [self.head(i) for i in inputs]

class RetinaNetHead_Row1(RetinaNetHead_Row3):
  def forward(self, inputs: List[Tensor]):
    for mod in self.head:
      if isinstance(mod, nn.BatchNorm2d):
        # for BN layer, normalize all inputs together
        shapes = [i.shape for i in inputs]
        spatial_sizes = [s[2] * s[3] for s in shapes]
        x = [i.flatten(2) for i in inputs]
        x = torch.cat(x, dim=2).unsqueeze(3)
        x = mod(x).split(spatial_sizes, dim=2)
        inputs = [i.view(s) for s, i in zip(shapes, x)]
      else:
        # for conv layer, apply it separately
        inputs = [mod(i) for i in inputs]
    return inputs

class RetinaNetHead_Row6:
  def __init__(self, num_conv, channel, num_features):
    # num_features: number of features coming from
    #               different FPN levels, e.g. 5
    heads = [[] for _ in range(num_levels)]
    for _ in range(num_conv):
      conv = nn.Conv2d(channel, channel, 3)
      for h in heads:
        # add a shared conv and a domain-specific BN
        h.extend([conv, nn.BatchNorm2d(channel)])
    self.heads = [nn.Sequential(*h) for h in heads]

  def forward(self, inputs: List[Tensor]):
    # end up with one head for each input
    return [head(i) for head, i in
             zip(self.heads, inputs)]
  \end{lstlisting}
\end{algorithm}

\subsection{Non-standard Interaction with Libraries}
\label{sec:interaction-with-libraries}

BatchNorm's unique properties as we discussed in this review are different
from most other deep learning primitives.
As a result, it interacts with the underlying deep learning systems
in its unique ways and if not paid attention to, can easily lead to bugs.
We now list a few types of issues caused by
such interactions,
and many of them echo with the specific examples given in \cite{brock2021high}.

\paragraph{Missing updates.}BatchNorm's parameters are not
modified by gradient updates, but often by EMA.
In declarative computation-graph execution libraries,
this requires explicit execution of these extra update operations during training.
In the BatchNorm layer implementation provided by TensorFlow v1~\cite{Abadi2016},
such operations are tracked using an \texttt{UPDATE\_OPS} operator collection.
Since models without BatchNorm do not require executing such operations,
users are likely to forget to execute them when using BatchNorm,
causing population statistics to not get updated.
TensorFlow \href{https://www.tensorflow.org/api_docs/python/tf/compat/v1/layers/batch_normalization}{provides a note}
to remind users of such misuse.

For the same reason, it is difficult to update the statistics if BatchNorm is used inside a symbolic
conditional branch such as \texttt{tf.cond}. Such problems were reported \href{https://github.com/tensorflow/tensorflow/issues/14699}{multiple times}.
To avoid these issues, Tensorpack \cite{wu2016tensorpack} and the now deprecated
\texttt{tf.contrib.layers} provide options that allow
BatchNorm to update its statistics during forward, instead of as extra operations.
This may however lead to the issue of unintended updates if not used carefully.

\paragraph{Unintended updates.}Imperative deep learning libraries like PyTorch~\cite{Paszke2017}
have to update the BatchNorm parameters as part of the forward computation.
This ensures updates will be executed, but it may cause unintended updates:
when the model is in ``training mode'', simply running forward passes will update
the model parameters, even without any back-propagation.
We noticed some open source code performed training on
validation set in this way, either accidentally or intentionally.

\paragraph{Empty inputs.}When a batch contains 0 sample (common in certain models such as R-CNN),
BatchNorm challenges the underlying library more. It may require
control flow support to not update EMA as the statistics are undefined.
When using SyncBN, handling empty inputs
and control flow without communication deadlocks is also a challenge.

\paragraph{Freezing.}Some high-level training libraries introduce the concept of ``freezing''
layers or setting layers to ``untrainable'' state.
Such concept is ambiguous for BatchNorm because it is not trained
by gradient descent like other layers, and therefore can cause confusions or misuse.
For example, \href{https://github.com/keras-team/keras/pull/9965}{these discussions}
unveil the widespread misuse in Keras.

\paragraph{Numerical precision.} Division is an operation in normalization
that's rarely seen in other deep learning primitives.
Division by a small number can easily introduce numerical instability,
which is likely to happen if inputs can be almost identical
in certain applications
(e.g. sparse features, reinforcement learning using simulator inputs).
\cite{brock2018large} increases $\epsilon$ in the denominator to counter
such issues.

Large reduction as part of normalization is also a potential source of instability.
In mixed precision training~\cite{micikevicius2017mixed}, such reduction often has to
be performed in full precision due to the limited range of half precision.

\paragraph{Batch splitting.}Training systems based on pipeline parallelism prefer splitting mini-batch
into smaller micro-batches for efficient pipelining~\cite{huang2018gpipe}.
Pipelining and batch splitting are also seen in custom training hardwares
with limited RAM~\cite{lym2018mini}.
Though splitting is an equivalent transform for other layers, it
affects the computation of BatchNorm and complicates the design of such systems.

{\small
\bibliographystyle{ieee_fullname}
\bibliography{batchnorm}

\begin{thebibliography}{10}\itemsep=-1pt

\bibitem{fvcore}
{fvcore} library.
\newblock \url{https://github.com/facebookresearch/fvcore/}.

\bibitem{tensorflow_od_api}
{TensorFlow} object detection {API}.
\newblock
  \url{https://github.com/tensorflow/models/tree/master/research/object_detection}.

\bibitem{Abadi2016}
Mart{\'\i}n Abadi, Paul Barham, Jianmin Chen, Zhifeng Chen, Andy Davis, Jeffrey
  Dean, Matthieu Devin, Sanjay Ghemawat, Geoffrey Irving, Michael Isard, et~al.
\newblock Tensorflow: A system for large-scale machine learning.
\newblock In {\em Operating Systems Design and Implementation (OSDI)}, 2016.

\bibitem{awais2020towards}
Muhammad Awais, Fahad Shamshad, and Sung-Ho Bae.
\newblock Towards an adversarially robust normalization approach.
\newblock {\em arXiv preprint arXiv:2006.11007}, 2020.

\bibitem{Ba2016}
Jimmy~Lei Ba, Jamie~Ryan Kiros, and Geoffrey~E Hinton.
\newblock Layer normalization.
\newblock {\em arXiv:1607.06450}, 2016.

\bibitem{benz2021revisiting}
Philipp Benz, Chaoning Zhang, Adil Karjauv, and In~So Kweon.
\newblock Revisiting batch normalization for improving corruption robustness.
\newblock In {\em Proceedings of the IEEE/CVF Winter Conference on Applications
  of Computer Vision}, pages 494--503, 2021.

\bibitem{brock2021high}
Andrew Brock, Soham De, Samuel~L Smith, and Karen Simonyan.
\newblock High-performance large-scale image recognition without normalization.
\newblock {\em arXiv preprint arXiv:2102.06171}, 2021.

\bibitem{brock2018large}
Andrew Brock, Jeff Donahue, and Karen Simonyan.
\newblock Large scale {GAN} training for high fidelity natural image synthesis.
\newblock In {\em ICLR}, 2019.

\bibitem{cai2021exponential}
Zhaowei Cai, Avinash Ravichandran, Subhransu Maji, Charless Fowlkes, Zhuowen
  Tu, and Stefano Soatto.
\newblock Exponential moving average normalization for self-supervised and
  semi-supervised learning.
\newblock {\em arXiv preprint arXiv:2101.08482}, 2021.

\bibitem{cariucci2017autodial}
Fabio~Maria Cariucci, Lorenzo Porzi, Barbara Caputo, Elisa Ricci, and
  Samuel~Rota Bulo.
\newblock Autodial: Automatic domain alignment layers.
\newblock In {\em ICCV}, 2017.

\bibitem{chang2019domain}
Woong-Gi Chang, Tackgeun You, Seonguk Seo, Suha Kwak, and Bohyung Han.
\newblock Domain-specific batch normalization for unsupervised domain
  adaptation.
\newblock In {\em CVPR}, 2019.

\bibitem{chen2020simple}
Ting Chen, Simon Kornblith, Mohammad Norouzi, and Geoffrey Hinton.
\newblock A simple framework for contrastive learning of visual
  representations.
\newblock In {\em ICML}, pages 1597--1607. PMLR, 2020.

\bibitem{sample_variance}
Eungchum Cho and Moon~Jung Cho.
\newblock Variance of sample variance.
\newblock In {\em Survey Research Methods Section}, 2008.

\bibitem{Deng2009}
Jia Deng, Wei Dong, Richard Socher, Li-Jia Li, Kai Li, and Li Fei-Fei.
\newblock {ImageNet: A large-scale hierarchical image database}.
\newblock In {\em CVPR}, 2009.

\bibitem{ghiasi2018dropblock}
Golnaz Ghiasi, Tsung-Yi Lin, and Quoc~V Le.
\newblock Dropblock: a regularization method for convolutional networks.
\newblock In {\em NeurIPS}, 2018.

\bibitem{Girshick2015}
Ross Girshick.
\newblock {Fast R-CNN}.
\newblock In {\em ICCV}, 2015.

\bibitem{Girshick2014}
Ross Girshick, Jeff Donahue, Trevor Darrell, and Jitendra Malik.
\newblock Rich feature hierarchies for accurate object detection and semantic
  segmentation.
\newblock In {\em CVPR}, 2014.

\bibitem{Goyal2017}
Priya Goyal, Piotr Doll{\'a}r, Ross Girshick, Pieter Noordhuis, Lukasz
  Wesolowski, Aapo Kyrola, Andrew Tulloch, Yangqing Jia, and Kaiming He.
\newblock Accurate, large minibatch {SGD}: Training {ImageNet} in 1 hour.
\newblock {\em arXiv:1706.02677}, 2017.

\bibitem{grill2020bootstrap}
Jean-Bastien Grill, Florian Strub, Florent Altch\'{e}, Corentin Tallec, Pierre
  Richemond, Elena Buchatskaya, Carl Doersch, Bernardo Avila~Pires, Zhaohan
  Guo, Mohammad Gheshlaghi~Azar, Bilal Piot, koray kavukcuoglu, Remi Munos, and
  Michal Valko.
\newblock Bootstrap your own latent - a new approach to self-supervised
  learning.
\newblock In {\em NeurIPS}, 2020.

\bibitem{guo2018double}
Yong Guo, Qingyao Wu, Chaorui Deng, Jian Chen, and Mingkui Tan.
\newblock Double forward propagation for memorized batch normalization.
\newblock In {\em Proceedings of the AAAI Conference on Artificial
  Intelligence}, volume~32, 2018.

\bibitem{he2019momentum}
Kaiming He, Haoqi Fan, Yuxin Wu, Saining Xie, and Ross Girshick.
\newblock Momentum contrast for unsupervised visual representation learning.
\newblock In {\em CVPR}, 2020.

\bibitem{he2019rethinking}
Kaiming He, Ross Girshick, and Piotr Doll{\'a}r.
\newblock Rethinking imagenet pre-training.
\newblock In {\em ICCV}, 2019.

\bibitem{He2017}
Kaiming He, Georgia Gkioxari, Piotr Doll{\'a}r, and Ross Girshick.
\newblock {Mask R-CNN}.
\newblock In {\em ICCV}, 2017.

\bibitem{He2016}
Kaiming He, Xiangyu Zhang, Shaoqing Ren, and Jian Sun.
\newblock Deep residual learning for image recognition.
\newblock In {\em CVPR}, 2016.

\bibitem{Henaff2019}
Olivier Henaff.
\newblock Data-efficient image recognition with contrastive predictive coding.
\newblock In {\em ICML}, 2020.

\bibitem{hendrycks2019benchmarking}
Dan Hendrycks and Thomas Dietterich.
\newblock Benchmarking neural network robustness to common corruptions and
  perturbations.
\newblock In {\em ICLR}, 2019.

\bibitem{hoffer2017train}
Elad Hoffer, Itay Hubara, and Daniel Soudry.
\newblock Train longer, generalize better: closing the generalization gap in
  large batch training of neural networks.
\newblock In {\em NeurIPS}, pages 1731--1741, 2017.

\bibitem{Hu_2018_CVPR}
Jie Hu, Li Shen, and Gang Sun.
\newblock Squeeze-and-excitation networks.
\newblock In {\em CVPR}, 2018.

\bibitem{Huang2016deep}
Gao Huang, Yu Sun, Zhuang Liu, Daniel Sedra, and Kilian~Q Weinberger.
\newblock Deep networks with stochastic depth.
\newblock In {\em ECCV}, 2016.

\bibitem{huang2018gpipe}
Yanping Huang, Youlong Cheng, Ankur Bapna, Orhan Firat, Dehao Chen, Mia Chen,
  HyoukJoong Lee, Jiquan Ngiam, Quoc~V Le, Yonghui Wu, and zhifeng Chen.
\newblock Gpipe: Efficient training of giant neural networks using pipeline
  parallelism.
\newblock In {\em NeurIPS}, 2019.

\bibitem{hubara2020improving}
Itay Hubara, Yury Nahshan, Yair Hanani, Ron Banner, and Daniel Soudry.
\newblock Improving post training neural quantization: Layer-wise calibration
  and integer programming.
\newblock {\em arXiv preprint arXiv:2006.10518}, 2020.

\bibitem{Ioffe2017}
Sergey Ioffe.
\newblock Batch renormalization: Towards reducing minibatch dependence in
  batch-normalized models.
\newblock In {\em NeurIPS}, 2017.

\bibitem{Ioffe2015}
Sergey Ioffe and Christian Szegedy.
\newblock Batch normalization: Accelerating deep network training by reducing
  internal covariate shift.
\newblock In {\em ICML}, 2015.

\bibitem{Isola2017}
Phillip Isola, Jun-Yan Zhu, Tinghui Zhou, and Alexei~A Efros.
\newblock Image-to-image translation with conditional adversarial networks.
\newblock In {\em CVPR}, 2017.

\bibitem{Pavel_SWA}
Pavel Izmailov, Dmitrii Podoprikhin, Timur Garipov, Dmitry Vetrov, and {Andrew
  Gordon} Wilson.
\newblock Averaging weights leads to wider optima and better generalization.
\newblock In {\em 34th Conference on Uncertainty in Artificial Intelligence
  2018, UAI 2018}, 2018.

\bibitem{Jia2014}
Yangqing Jia, Evan Shelhamer, Jeff Donahue, Sergey Karayev, Jonathan Long, Ross
  Girshick, Sergio Guadarrama, and Trevor Darrell.
\newblock Caffe: Convolutional architecture for fast feature embedding.
\newblock In {\em Proceedings of the 22nd ACM international conference on
  Multimedia}, pages 675--678, 2014.

\bibitem{johnson2018image}
Justin Johnson, Agrim Gupta, and Li Fei-Fei.
\newblock Image generation from scene graphs.
\newblock In {\em CVPR}, 2018.

\bibitem{alex2019big}
Alexander Kolesnikov, Lucas Beyer, Xiaohua Zhai, Joan Puigcerver, Jessica Yung,
  Sylvain Gelly, and Neil Houlsby.
\newblock Big transfer (bit): General visual representation learning, 2019.

\bibitem{krishnamoorthi2018quantizing}
Raghuraman Krishnamoorthi.
\newblock Quantizing deep convolutional networks for efficient inference: A
  whitepaper.
\newblock {\em arXiv preprint arXiv:1806.08342}, 2018.

\bibitem{li2020feature}
Boyi Li, Felix Wu, Ser-Nam Lim, Serge Belongie, and Kilian~Q. Weinberger.
\newblock On feature normalization and data augmentation, 2020.

\bibitem{Li_2019_CVPR}
Rundong Li, Yan Wang, Feng Liang, Hongwei Qin, Junjie Yan, and Rui Fan.
\newblock Fully quantized network for object detection.
\newblock In {\em CVPR}, 2019.

\bibitem{Li_dropout_bn}
Xiang Li, Shuo Chen, Xiaolin Hu, and Jian Yang.
\newblock Understanding the disharmony between dropout and batch normalization
  by variance shift.
\newblock In {\em CVPR}, 2019.

\bibitem{li2016revisiting}
Yanghao Li, Naiyan Wang, Jianping Shi, Jiaying Liu, and Xiaodi Hou.
\newblock Revisiting batch normalization for practical domain adaptation.
\newblock {\em arXiv preprint arXiv:1603.04779}, 2016.

\bibitem{li2021momentum}
Zeming Li, Songtao Liu, and Jian Sun.
\newblock Momentum\^{} 2 teacher: Momentum teacher with momentum statistics for
  self-supervised learning.
\newblock {\em arXiv preprint arXiv:2101.07525}, 2021.

\bibitem{Lin2017}
Tsung-Yi Lin, Piotr Doll{\'a}r, Ross Girshick, Kaiming He, Bharath Hariharan,
  and Serge Belongie.
\newblock Feature pyramid networks for object detection.
\newblock In {\em CVPR}, 2017.

\bibitem{Lin2017a}
Tsung-Yi Lin, Priya Goyal, Ross Girshick, Kaiming He, and Piotr Doll{\'a}r.
\newblock Focal loss for dense object detection.
\newblock In {\em ICCV}, 2017.

\bibitem{liu2020evolving}
Hanxiao Liu, Andrew Brock, Karen Simonyan, and Quoc~V Le.
\newblock Evolving normalization-activation layers.
\newblock {\em NeurIPS}, 2020.

\bibitem{lym2018mini}
Sangkug Lym, Armand Behroozi, Wei Wen, Ge Li, Yongkee Kwon, and Mattan Erez.
\newblock Mini-batch serialization: Cnn training with inter-layer data reuse.
\newblock In {\em Proceedings of Machine Learning and Systems}, 2019.

\bibitem{micikevicius2017mixed}
Paulius Micikevicius, Sharan Narang, Jonah Alben, Gregory Diamos, Erich Elsen,
  David Garcia, Boris Ginsburg, Michael Houston, Oleksii Kuchaiev, Ganesh
  Venkatesh, et~al.
\newblock Mixed precision training.
\newblock {\em arXiv preprint arXiv:1710.03740}, 2017.

\bibitem{musgrave2020metric}
Kevin Musgrave, Serge Belongie, and Ser-Nam Lim.
\newblock A metric learning reality check.
\newblock {\em arXiv preprint arXiv:2003.08505}, 2020.

\bibitem{nado2020evaluating}
Zachary Nado, Shreyas Padhy, D Sculley, Alexander D'Amour, Balaji
  Lakshminarayanan, and Jasper Snoek.
\newblock Evaluating prediction-time batch normalization for robustness under
  covariate shift.
\newblock {\em arXiv preprint arXiv:2006.10963}, 2020.

\bibitem{yanninterview}
Computer~Vision News.
\newblock Exclusive interview with {Yann LeCun}.
\newblock \url{https://www.rsipvision.com/ComputerVisionNews-2018November/9/},
  page 9, 2018.

\bibitem{nichol2018first}
Alex Nichol, Joshua Achiam, and John Schulman.
\newblock On first-order meta-learning algorithms.
\newblock {\em arXiv preprint arXiv:1803.02999}, 2018.

\bibitem{cifar10in1m}
David Page.
\newblock How to train your resnet 8: Bag of tricks.
\newblock \url{https://myrtle.ai/how-to-train-your-resnet-8-bag-of-tricks/},
  2019.

\bibitem{pan2018two}
Xingang Pan, Ping Luo, Jianping Shi, and Xiaoou Tang.
\newblock Two at once: Enhancing learning and generalization capacities via
  ibn-net.
\newblock In {\em ECCV}, 2018.

\bibitem{Paszke2017}
Adam Paszke, Sam Gross, Soumith Chintala, Gregory Chanan, Edward Yang, Zachary
  DeVito, Zeming Lin, Alban Desmaison, Luca Antiga, and Adam Lerer.
\newblock Automatic differentiation in pytorch.
\newblock 2017.

\bibitem{Peng2018}
Chao Peng, Tete Xiao, Zeming Li, Yuning Jiang, Xiangyu Zhang, Kai Jia, Gang Yu,
  and Jian Sun.
\newblock {MegDet}: A large mini-batch object detector.
\newblock In {\em CVPR}, 2018.

\bibitem{Radosavovic_2020_CVPR}
Ilija Radosavovic, Raj~Prateek Kosaraju, Ross Girshick, Kaiming He, and Piotr
  Dollar.
\newblock Designing network design spaces.
\newblock In {\em CVPR}, June 2020.

\bibitem{Ren2015}
Shaoqing Ren, Kaiming He, Ross Girshick, and Jian Sun.
\newblock {Faster R-CNN}: Towards real-time object detection with region
  proposal networks.
\newblock In {\em NeurIPS}, 2015.

\bibitem{namedtensor}
Alexander Rush.
\newblock Tensor considered harmful.
\newblock \url{http://nlp.seas.harvard.edu/NamedTensor}, 2019.

\bibitem{salimans2016improved}
Tim Salimans, Ian Goodfellow, Wojciech Zaremba, Vicki Cheung, Alec Radford, and
  Xi Chen.
\newblock Improved techniques for training {GANs}.
\newblock In {\em NeurIPS}, 2016.

\bibitem{schneider2020improving}
Steffen Schneider, Evgenia Rusak, Luisa Eck, Oliver Bringmann, Wieland Brendel,
  and Matthias Bethge.
\newblock Improving robustness against common corruptions by covariate shift
  adaptation.
\newblock {\em NeurIPS}, 2020.

\bibitem{shen2020powernorm}
Sheng Shen, Zhewei Yao, Amir Gholami, Michael Mahoney, and Kurt Keutzer.
\newblock Powernorm: Rethinking batch normalization in transformers.
\newblock In {\em ICML}, 2020.

\bibitem{shomron2020post}
Gil Shomron and Uri Weiser.
\newblock Post-training batchnorm recalibration.
\newblock {\em arXiv preprint arXiv:2010.05625}, 2020.

\bibitem{singh2019filter}
Saurabh Singh and Shankar Krishnan.
\newblock Filter response normalization layer: Eliminating batch dependence in
  the training of deep neural networks.
\newblock {\em CVPR}, 2019.

\bibitem{singh2019evalnorm}
Saurabh Singh and Abhinav Shrivastava.
\newblock Evalnorm: Estimating batch normalization statistics for evaluation.
\newblock In {\em ICCV}, 2019.

\bibitem{smith2017don}
Samuel~L Smith, Pieter-Jan Kindermans, Chris Ying, and Quoc~V Le.
\newblock Don't decay the learning rate, increase the batch size.
\newblock {\em ICLR}, 2018.

\bibitem{Srivastava2014}
Nitish Srivastava, Geoffrey Hinton, Alex Krizhevsky, Ilya Sutskever, and Ruslan
  Salakhutdinov.
\newblock Dropout: A simple way to prevent neural networks from overfitting.
\newblock {\em The Journal of Machine Learning Research}, pages 1929--1958,
  2014.

\bibitem{sun2019hybrid}
Xiao Sun, Jungwook Choi, Chia-Yu Chen, Naigang Wang, Swagath Venkataramani,
  Vijayalakshmi~Viji Srinivasan, Xiaodong Cui, Wei Zhang, and Kailash
  Gopalakrishnan.
\newblock Hybrid 8-bit floating point ({HFP8}) training and inference for deep
  neural networks.
\newblock {\em NeurIPS}, 2019.

\bibitem{pmlr-v97-tan19a}
Mingxing Tan and Quoc Le.
\newblock {E}fficient{N}et: Rethinking model scaling for convolutional neural
  networks.
\newblock In {\em ICML}, 2019.

\bibitem{tian2019opengo}
Yuandong Tian, Jerry Ma, Qucheng Gong, Shubho Sengupta, Zhuoyuan Chen, James
  Pinkerton, and Larry Zitnick.
\newblock {ELF OpenGo}: an analysis and open reimplementation of alphazero.
\newblock In {\em ICML}, 2019.

\bibitem{Ulyanov2016}
Dmitry Ulyanov, Andrea Vedaldi, and Victor Lempitsky.
\newblock Instance normalization: The missing ingredient for fast stylization.
\newblock {\em arXiv:1607.08022}, 2016.

\bibitem{wang2019transferable}
Ximei Wang, Ying Jin, Mingsheng Long, Jianmin Wang, and Michael~I Jordan.
\newblock Transferable normalization: Towards improving transferability of deep
  neural networks.
\newblock In {\em NeurIPS}, 2019.

\bibitem{wu2019multigrid}
Chao-Yuan Wu, Ross Girshick, Kaiming He, Christoph Feichtenhofer, and Philipp
  Kr{\"a}henb{\"u}hl.
\newblock A multigrid method for efficiently training video models.
\newblock {\em CVPR}, 2020.

\bibitem{wu2016tensorpack}
Yuxin Wu et~al.
\newblock Tensorpack.
\newblock \url{https://github.com/tensorpack/}, 2016.

\bibitem{Wu2018}
Yuxin Wu and Kaiming He.
\newblock Group normalization.
\newblock In {\em ECCV}, 2018.

\bibitem{wu2019detectron2}
Yuxin Wu, Alexander Kirillov, Francisco Massa, Wan-Yen Lo, and Ross Girshick.
\newblock Detectron2.
\newblock \url{https://github.com/facebookresearch/detectron2}, 2019.

\bibitem{xie2019adversarial}
Cihang Xie, Mingxing Tan, Boqing Gong, Jiang Wang, Alan Yuille, and Quoc~V Le.
\newblock Adversarial examples improve image recognition.
\newblock {\em CVPR}, 2020.

\bibitem{xie2019intriguing}
Cihang Xie and Alan Yuille.
\newblock Intriguing properties of adversarial training.
\newblock {\em ICLR}, 2019.

\bibitem{yan2020towards}
Junjie Yan, Ruosi Wan, Xiangyu Zhang, Wei Zhang, Yichen Wei, and Jian Sun.
\newblock Towards stabilizing batch statistics in backward propagation of batch
  normalization.
\newblock {\em ICLR}, 2020.

\bibitem{yao2020cross}
Zhuliang Yao, Yue Cao, Shuxin Zheng, Gao Huang, and Stephen Lin.
\newblock Cross-iteration batch normalization.
\newblock {\em arXiv preprint arXiv:2002.05712}, 2020.

\bibitem{yun2019cutmix}
Sangdoo Yun, Dongyoon Han, Seong~Joon Oh, Sanghyuk Chun, Junsuk Choe, and
  Youngjoon Yoo.
\newblock Cutmix: Regularization strategy to train strong classifiers with
  localizable features.
\newblock In {\em ICCV}, 2019.

\bibitem{zajkac2019split}
Micha{\l} Zaj{\k{a}}c, Konrad {\.Z}o{\l}na, and Stanis{\l}aw Jastrz{\k{e}}bski.
\newblock Split batch normalization: Improving semi-supervised learning under
  domain shift.
\newblock {\em arXiv preprint arXiv:1904.03515}, 2019.

\bibitem{zhang2018mixup}
Hongyi Zhang, Moustapha Cisse, Yann~N. Dauphin, and David Lopez-Paz.
\newblock mixup: Beyond empirical risk minimization.
\newblock In {\em ICLR}, 2018.

\bibitem{Zhang_2018_CVPR}
Hang Zhang, Kristin Dana, Jianping Shi, Zhongyue Zhang, Xiaogang Wang, Ambrish
  Tyagi, and Amit Agrawal.
\newblock Context encoding for semantic segmentation.
\newblock In {\em CVPR}, June 2018.

\end{thebibliography}
}

%\end{document}

\newpage

\setcounter{page}{1}

% reset ruler
\newcount\cvprrulercount

\end{document}